\documentclass[journal,compsoc]{IEEEtran}
\usepackage{textcomp}
\usepackage{stfloats}
\usepackage{verbatim}

\usepackage{tabularx,booktabs}
\usepackage{color}
\usepackage{tabularray}
\UseTblrLibrary{diagbox}
\usepackage{ifpdf}
\ifpdf
\else
\fi

\usepackage{cite}

\ifCLASSINFOpdf
\usepackage[pdftex]{graphicx}
\else
\fi

\usepackage{amsmath}
\usepackage{amsfonts}
\usepackage{amssymb}
\usepackage{extarrows}

\makeatletter
\newif\if@restonecol
\makeatother

\usepackage[linesnumbered,ruled,lined]{algorithm2e}

\usepackage{array}
\usepackage{multirow}
\usepackage{makecell,rotating}       
\usepackage{float}

\usepackage{tikz}
\usetikzlibrary{positioning, shapes.geometric}
\usepackage{pgfplots}

\ifCLASSOPTIONcompsoc
\usepackage[caption=false,font=normalsize,labelfont=sf,textfont=sf]{subfig}
\else
\usepackage[caption=false,font=footnotesize]{subfig}
\fi

\usepackage{url}

\makeatletter
\let\NAT@parse\undefined
\makeatother
\usepackage[colorlinks,linkcolor=blue,anchorcolor=blue,citecolor=blue]{hyperref}
\begin{document}

\title{Random Graph Set and Evidence Pattern Reasoning Model}

\author{
	Tianxiang Zhan,
	Zhen Li,
	Yong Deng

	\thanks{This work was supported in part by the National Natural Science Foundation of China under Grant 62373078. }
	\thanks{Tianxiang Zhan is with the Institute of Fundamental and Frontier Science, University of Electronic Science and Technology of China, Chengdu 610054, China (e-mail: zhantianxianguestc@hotmail.com). }
	\thanks{Zhen Li is with the China Mobile Information Technology Center, Beijing 100029, China (e-mail: zhen.li@pku.edu.cn).}
	\thanks{Yong Deng is with the Institute of Fundamental and Frontier Science, University of Electronic Science and Technology of China, Chengdu 610054, China, and also with the School of Medicine, Vanderbilt University, Nashville, TN 37240 USA (e-mail: dengentropy@uestc.edu.cn).}
	\thanks{Corresponding author: \textit{Li Zhen} and \textit{Yong Deng}}
}

\mark{JOURNAL OF LATEX CLASS FILES, VOL. 14, NO. 8, AUGUST 2015}


\IEEEtitleabstractindextext{%
\begin{abstract}
Evidence theory is widely used in decision making and reasoning systems. In previous research, Transferable Belief Model (TBM) is a commonly used evidential decision making model, but TBM is a non-preference model.  In order to better fit the decision making goals, the Evidence Pattern Reasoning Model (EPRM) is proposed. By defining pattern operators and decision making operators, corresponding preferences can be set for different tasks. Random Permutation Set (RPS) expands order information for evidence theory. It is hard for RPS to characterize the complex relationship between samples such as cycling, paralleling relationships. Therefore, Random Graph Set (RGS) were proposed to model complex relationships and represent more event types.  In order to illustrate the significance of RGS and EPRM, an experiment of aircraft velocity ranking was designed and 10,000 cases were simulated. The implementation of  EPRM called Conflict Resolution Decision optimized 18.17\% of the cases compared to Mean Velocity Decision, effectively improving the aircraft velocity ranking. EPRM provides a unified solution for evidence-based decision making.
\end{abstract}

\begin{IEEEkeywords}
Decision Making, Evidential Reasoning, Pattern Recognition, Graph, Engineering
\end{IEEEkeywords}
}

\maketitle

\section{Introduction}
\IEEEPARstart{E}{vidence} theory is a theory for handling uncertainty, first proposed by Dempster and Shafer in 1976 which is also called Dempster-Shafer Theory (DST) \cite{dempster2008upper, shafer1976mathematical}. It is worth noting that DST is based on mass function, which has weaker constraints compared to probability and is a generalization of probability. As a method of information fusion and reasoning, DST is widely used in uncertainty measurement \cite{deng2021measuring}, optimization \cite{li2022new}, decision making \cite{liu2021consensus}, pattern recognition \cite{xiao2022generalized,xiao2022negation} and so on \cite{zhu2021fuzzy, zhao2023dpt}. In order to measure the uncertainty of evidence, Deng proposed Deng entropy and information volume of mass function \cite{deng2016deng, deng2020information}. Deng entropy  and information volume of mass function contributes to nonlinear systems \cite{qiang2022information, kharazmi2023deng, zhao2024linearity} and information measurement\cite{zhou2024improved}, also plays a role in time series analysis \cite{cui2022belief}.  Considering quantum theory, Pan et al. proposed quantum combination rules \cite{pan2023quantum}. In DST, the focal element in the event space focuses on the combination relationship of samples. In the real world, there are not only combination relationships but also permutation relationships, so Random Permutation Set (RPS) was proposed to solve the problem of containing order information \cite{deng2022random, deng2022maximum, chen2023entropy}. It is conceivable that RPS has expanded the application scenarios of DST \cite{chen2023distance, zhao2023information}. 

On the other hand, DST mainly plays a role in reasoning and decision making. In 1991, Smets and Kennes proposed the Transferable Belief Model (TBM), which provided a novel sight for the transfer of mass function in DST \cite{smets1994transferable}. Along with TBM, another major method is Pignistic Probability Transformation (PPT) which transforms mass function to probability mass function (PMF)\cite{smets2005decision}. Zhou et al. proposed Belief Evolution Network (BEN) which further explores the transferable mechanism \cite{zhou2022belief}. Evidential Reasoning (ER) is another representative of DST in reasoning which is proposed by Yang and Singh \cite{yang1994evidential}. Then, ER rule is extended from basic ER algorithm which involves the weight and reliability by Yang and Xu \cite{yang2013evidential}. The common Bayesian inference method is based on probability. Yang and Xu explained that Bayesian inference is a special case of ER in 2014 \cite{yang2014study}.

However, both previous TBM and RPS can be deeply optimized. First, the TBM model is a preference-free model. This means that the information in the entire decision making system has not changed and can be converted between the mass function and the corresponding PMF. This makes it difficult in engineering practice to adjust hyperparameters outside the decision making system more realistically, which is called preference. \textbf{For example, at the time of trading in quantitative trading, there are different decision making modes of high risk with high return and low risk with low return. } This preference needs to be decided by the user, so this article proposes an \textbf{Evidence Pattern Reasoning Model (EPRM)} to accommodate different preferences. In EPRM, three operators Basic Probability Assignment (BPA) Operator, \textbf{Pattern Operator (PO) and Decision Making Operator (DMO)} are abstractly designed to suit user preferences. BPA is a method that exists in previous work and is used for the generation of evidence. Subjective generation methods can be added to BPA. PO proposed in this article is generalized form of the combination operator in DST. And PO allows users to define a fusion method that conforms to actual patterns to act between focal elements of different evidence. DMO defines the decision making process generated by fused evidence, and is also an operator that can involves preferences. When the DMO has no preference, it degenerates into PPT. EPRM provides a unified framework for the reasoning process with preferences also allows the decision making system to accept external information. Only relevant functions need to be implemented to reduce workflow.

On the other hand, focal element representation also needs improvement. RPS introduces pure order, but RPS cannot directly represent combinations such as DST, and cannot represent complex order relationships such as cyclic relationships and parallel relationships. Therefore, Random Graph Set (RGS) was proposed to use graph to represent the relationship between samples. This means that focal elements in DST are represented by graphs, where nodes are composed of samples and edges can represent relationships between samples. The advantage of graphs is that they can directly represent relationships, reducing the difficulty of data representation. Another advantage is that the graph algorithms are quite complete, such as the intersection, union and composed operator of graphs. These operations are perfectly consistent with those of composition and ordering, since composition and ordering are special cases of graphs. It greatly reduces the difficulty of modeling complex relationships and engineering implementation by DST and RPS.

In order to vividly show the practical application of EPRM and RGS, this article provides a complex scenario of aircraft velocity sorting. A total of 10,000 cases were generated using stochastic process methods for both aircraft trajectories and sensor layouts. The dataset is entirely machine-generated and has been open sourced. The theoretical velocities of different aircraft are similar, while the actual flying velocity of the aircraft is set to have a large fluctuation range. As a result, the sensor will detect velocity data that is difficult to sort, and it is even possible to collect data that is completely opposite to the theoretical sorting. So this article implemented the EPRM framework, named the \textbf{Conflict Resolution Decision (CRD)} method. For reference, it chose to make a comparison based entirely on the logic of mean velocity measurement, and named it Mean Velocity Decision (MVD). \textbf{The experimental results show that CRD correctly optimized 18.17\% of the cases with MVD errors or conflicts in the entire sample.} At the same time, the effect of CVD will only be better than MRD, because CVD is optimized MRD, and MRD is the lower limit of CVD performance. Therefore, the contributions of this article are as follows:
\begin{itemize}
	\item \textbf{A unified evidential decision making/reasoning framework EPRM with preferences is proposed.
	\item RPS is further extended to RGS.}
	\item \textbf{EPRM and RGS optimize the process of engineering practice.}
	\item \textbf{This article provides an aircraft velocity sorting dataset.}
	\item \textbf{This article provides a multi-sensor collaborative velocity ranking algorithm CRD, and is an example of EPRM in engineering practice.}
\end{itemize}

This article is structured as follows. Section \ref{sec:pre} introduces the relevant basic theories. Section \ref{sec:EPRM} introduces the definition and implementation details of EPRM. Section \ref{sec:RGS} introduces the definition and implementation details of RGS. Section \ref{sec:air} shows a practical case of aircraft velocity ranking. Finally, Section \ref{sec:conclusion} summarizes the entire article.

\section{Preliminary} \label{sec:pre}

Sample Space $\Omega$ in Eq.\eqref{eq:ss} is composed of all possible base events $A_i$. The size of  sample space is $n$.

\begin{equation}
 \Omega = \{A_1,A_2,A_3,...,A_n\}
 \label{eq:ss}
\end{equation}

Mass function $m(\cdot)$ is a map in Eq.\eqref{eq:mass_function} where $\mathcal{E}$ is event space of $\Omega$. The constraint of $m(\cdot)$ is in Eq.\eqref{eq:mass_function_constraint}.

\begin{equation}
 m: \mathcal{E} \rightarrow [0,1]
 \label{eq:mass_function}
\end{equation}

\begin{equation}
\sum_{i \in S}m(i)=1 \quad and \quad m(i) \geqslant 0 \quad and \quad m(\emptyset)=0
 \label{eq:mass_function_constraint}
\end{equation}

\subsection{Dempster-Shafer Theory}
DST is also called evidence theory. Combination of $\Omega$ is considered by DST. In Eq.\eqref{eq:power_space}, $\mathcal{E}$ of evidence theory is power set $2^\Omega$ which contains all possible combination of $\Omega$.

\begin{equation}
	\begin{split}
		\mathcal{E} = 2^\Omega -\emptyset
		&= \{C\ |\ \forall C \subset \Omega \}\\
		&=\{\{A_1\},\{A_2\},...,\{A_n\},\{A_1,A_2\},...,\Omega\}
	\end{split}
 \label{eq:power_space}
\end{equation}

A piece of evidence is defined in Eq.\eqref{eq:evidence}.

\begin{equation}
	\begin{split}
		\varpi
		&=\{(C_i,m(C_i)\ |\ \forall i \in [1,2^n-1] \} \\
		&=\{(C_1, m(C_1)),(C_2, m(C_2)),...,(C_{2^n-1}, m(C_{2^n-1}))\}
	\end{split}
 \label{eq:evidence}
\end{equation}

Fused evidence $\varpi_\oplus$ of evidences $\varpi_1,...,\varpi_k$ is defined in Eq.\eqref{eq:evidence_combination} and Eq.\eqref{eq:evidence_combination_res} where $m_i(\cdot)$ is mass function of $i$-order evidence. Combination operator is noted as $\oplus$. $\oplus$ are commutative in Eq.\eqref{eq:evidence_combination_commutative} where $\varpi'$ is order-shuffled evidences.

\begin{equation}
		m(C) = \left\{
		\begin{aligned}
			&\frac{\sum_{\forall C_1 \cap C_2...\cap C_k = C}\prod_{i=1}^{k}m_i(C_{i})}{m(\emptyset)}\quad (C \neq \emptyset) \\
			&\sum_{\forall C_1 \cap C_2...\cap C_k = C}\prod_{i=1}^{k}m_i(C_{i})\quad (C = \emptyset)
		\end{aligned}
		\right.
		\label{eq:evidence_combination}
\end{equation}

\begin{equation}
	\begin{split}
		\varpi_\oplus =
		&\{(C_\theta, \frac{\sum_{\forall C_1 \cap C_2...\cap C_k = C_\theta}\prod_{i=1}^{k}m_i(C_{i})}{m(\emptyset)}) \\
		&|\ \forall \theta \in [1, 2^n-1]\}
	\end{split}
 \label{eq:evidence_combination_res}
\end{equation}

\begin{equation}
\varpi_\oplus =\varpi_1 \oplus \varpi_2 \oplus ...\oplus \varpi_k = \varpi_1' \oplus \varpi_2' \oplus ...\oplus \varpi_k'
\label{eq:evidence_combination_commutative}
\end{equation}

\subsection{Transferable Belief Model and Decision Making}
\begin{figure}[htbp]
	\centering
	\begin{tikzpicture}
		\node[circle, minimum width =30pt , minimum height =30pt ,draw=red] (ABC) at (0,0) {$A,B,C$};
		\node[circle, minimum width =30pt , minimum height =30pt ,draw=orange] (AB) at (-2,-2) {$A,B$};
		\node[circle, minimum width =30pt , minimum height =30pt ,draw=orange] (AC) at (0,-2) {$A,C$};
		\node[circle, minimum width =30pt , minimum height =30pt ,draw=orange] (BC) at (2,-2) {$B,C$};
		\node[circle, minimum width =30pt , minimum height =30pt ,draw=blue] (A) at (-3,-4) {$A$};
		\node[circle, minimum width =30pt , minimum height =30pt ,draw=blue] (B) at (0,-4) {$B$};
		\node[circle, minimum width =30pt , minimum height =30pt ,draw=blue] (C) at (3,-4) {$C$};
		\draw[->] (ABC)--(AB);
		\draw[->] (ABC)--(AC);
		\draw[->] (ABC)--(BC);
		\draw[->] (AB)--(A);
		\draw[->] (AB)--(B);
		\draw[->] (AC)--(A);
		\draw[->] (AC)--(C);
		\draw[->] (BC)--(B);
		\draw[->] (BC)--(C);
	\end{tikzpicture}
	\caption{Direction of transference in TBM (FOD $\Omega=\{A,B,C\}$)}
	\label{fig:tbm_direction}
\end{figure}
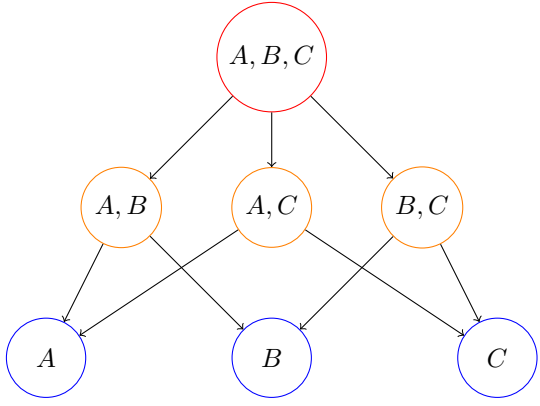

According to DST, TBM is proposed by Smets and Kennes. In TBM, assigned mass function of event $a$ transfers to mass function of further event $b, a \in b$ such as a case in Fig.\ref{fig:tbm_direction}. Open world is accepted by TBM, and mass function of $\emptyset$ is possible to be larger than $0$. In the process of decision making, PPT is necessary to transform basic probability assignment to PMF and is defined in Eq.\eqref{eq:ppt}.

\begin{equation}
	\begin{split}
		BetP(\theta) = \sum_{\theta \in A} \frac{m(A)}{|F| - m(\emptyset)}
	\end{split}
	\label{eq:ppt}
\end{equation}

\subsection{Random Permutation Set}
Event space is not limited to combination and permutation of $\Omega$ is considered by RPS. In Eq.\eqref{eq:permutation_event_space}, $\mathcal{E}$ in RPS is Permutation Event Space (PES) which contains all possible permutation of $\Omega$. And a piece of RPS is defined  in Eq.\eqref{eq:rps}. 

\begin{equation}
	\begin{split}
		\mathcal{E} &= PES(\Omega)\\
		&= \{p\ |\ \forall P\ is\ permutation\ of\ subset\ of\ \Omega \} -\emptyset \\
		&=\{p_{ij}|i=0,...,n;j=1,...,P(n,i)\} -\emptyset \\
		&=\left\{(A_1),(A_2),...,(A_n),(A_1,A_2),(A_2, A_1),...,        \right. \\
		&\left.(A_{n-1},A_n),(A_n,A_{n-1}),...,(A_1,A_2,...,A_N)),  \right. \\
		&\left. (A_n,A_{n-1},...,A_1) \right\}
	\end{split}
 \label{eq:permutation_event_space}
\end{equation}

\begin{equation}
	\begin{split}
		RPS=
		&\left\{(p_{ij},m(p_{ij}))\ |\ i=0,...,n;j=1, \right. \\
		&\left. ...,P(n,i) \right\}
	\end{split}
 \label{eq:rps}
\end{equation}

Due to the order information of event, the union operator $\cap$ loses commutative properties and transforms to left combination operator $\mathop{\cap}\limits^{\leftarrow}$ and right combination operator $\mathop{\cap}\limits^{\rightarrow}$ in Eq.\eqref{eq:rps_lr_union}. The direction of the arrow indicates the sequence based on the event being pointed to and $(a//b)$ stands for removing all elements in $b$ from order $a$. 

\begin{equation}
	\begin{split}
	C_1 \mathop{\cap}\limits^{\rightarrow} C_2=C_2 \mathop{\cap}\limits^{\leftarrow} C_1 = C_2//(C_1\cap C_2)
	\end{split}
 \label{eq:rps_lr_union}
\end{equation}

Right fused RPS $\varpi_\oplus$ of RPS $\varpi_1,...,\varpi_k$ is proposed by Deng in Eq.\eqref{eq:rps_combination} and Eq.\eqref{eq:rps_combination_res} where $m_i(\cdot)$ is mass function of $i$-order RPS.

\begin{equation}
	m(C) = \left\{
	\begin{aligned}
		&\frac{\sum_{\forall  \mathop{\bigcap}\limits^{\rightarrow}(C_1,C_2,...,C_k)= C}\prod_{i= 1}^{k}m_i(C_{i})}{m(\emptyset)} \quad (C \neq \emptyset) \\
		&\sum_{\forall  \mathop{\bigcap}\limits^{\rightarrow}(C_1,C_2,...,C_k)= C}\prod_{i= 1}^{k}m_i(C_{i})\quad (C = \emptyset)
	\end{aligned}
	\right.
	\label{eq:rps_combination}
\end{equation}

\begin{equation}
	\begin{split}
	\varpi_{\mathop{\cap}\limits^{\rightarrow}} = 
	&\{(C, \frac{\sum_{\forall  \mathop{\bigcap}\limits^{\rightarrow}(C_1,C_2,...,C_k)= C}\prod_{i= 1}^{k}m_i(C_{i})}{m(\emptyset)}) \\
	&|\ \forall C \in PES(\Omega)\}
	\end{split}
	\label{eq:rps_combination_res}
\end{equation}

\section{Evidence Pattern Reasoning Model}\label{sec:EPRM}

\begin{figure*}[htbp]
	\centering
	\includegraphics[width=0.95\linewidth]{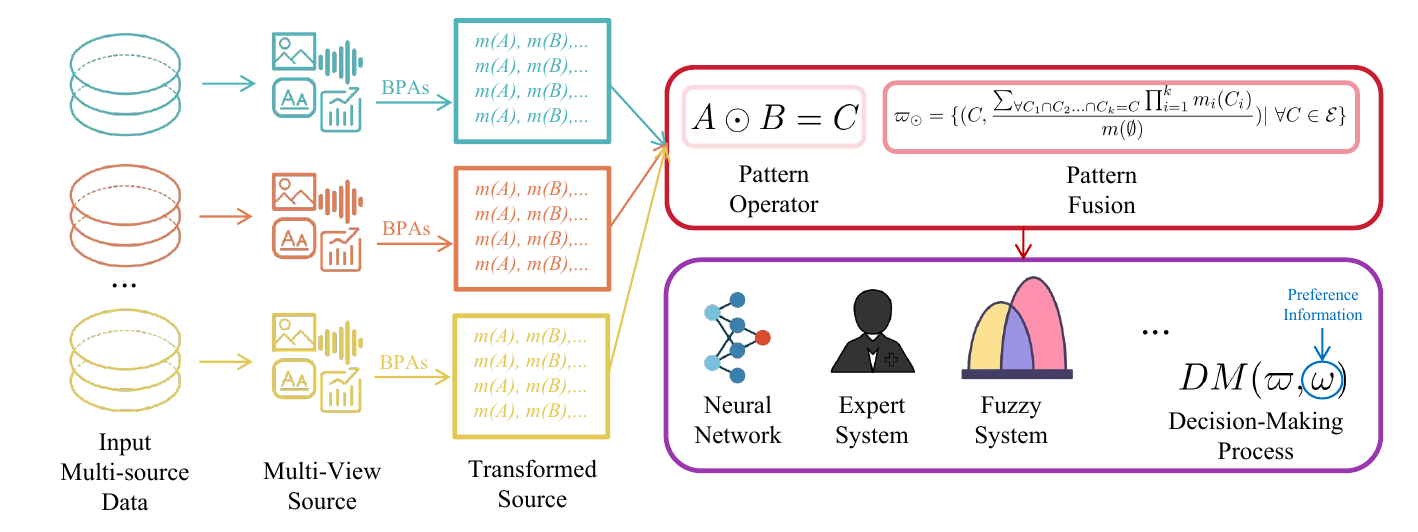}
	\caption{Process flow of belief pattern reasoning model}
	\label{fig:bpim}
\end{figure*}

EPRM is a reasoning framework based on mass function and has ability to fuse different type sources. The structure of EPRM is in Fig.\ref{fig:bpim} and relative symbols is listed in Tab.\ref{tab:bpim_sym}.

\begin{table}[htbp]
	\centering
	\caption{Symbols in EPRM}
	\resizebox{0.75\linewidth}{!}{
		\setlength{\tabcolsep}{12pt}
			\begin{tabular}{cc}
			\hline
			Variable	&  Symbol  \\ \hline
			Sample space & $\Omega$   \\
			Event space & $\mathcal{E}$   \\
			Mass function & $m(\cdot)$ \\
			Evidential Source & $\varpi$ \\
			Pattern operator & $\odot$   \\
			Cartesian product&  $\otimes$  \\ 
			Decision making function & $DM(\cdot)$ \\
			Preference parameter & $\omega$ \\ \hline
			\end{tabular}}
	\label{tab:bpim_sym}
\end{table}

\subsection{Confirming Sample Space And Event Space}
Sample space is a set contained all possible \textbf{atomic samples}.  Event space is a set contained \textbf{all possible relations} which is called as event. \textbf{In EPRM, event space does not contain empty set because empty set is a middle state of fusion and empty set does not express any relations between existed samples. }

\begin{figure}[htbp]
	\centering
	\includegraphics[width=0.9\linewidth]{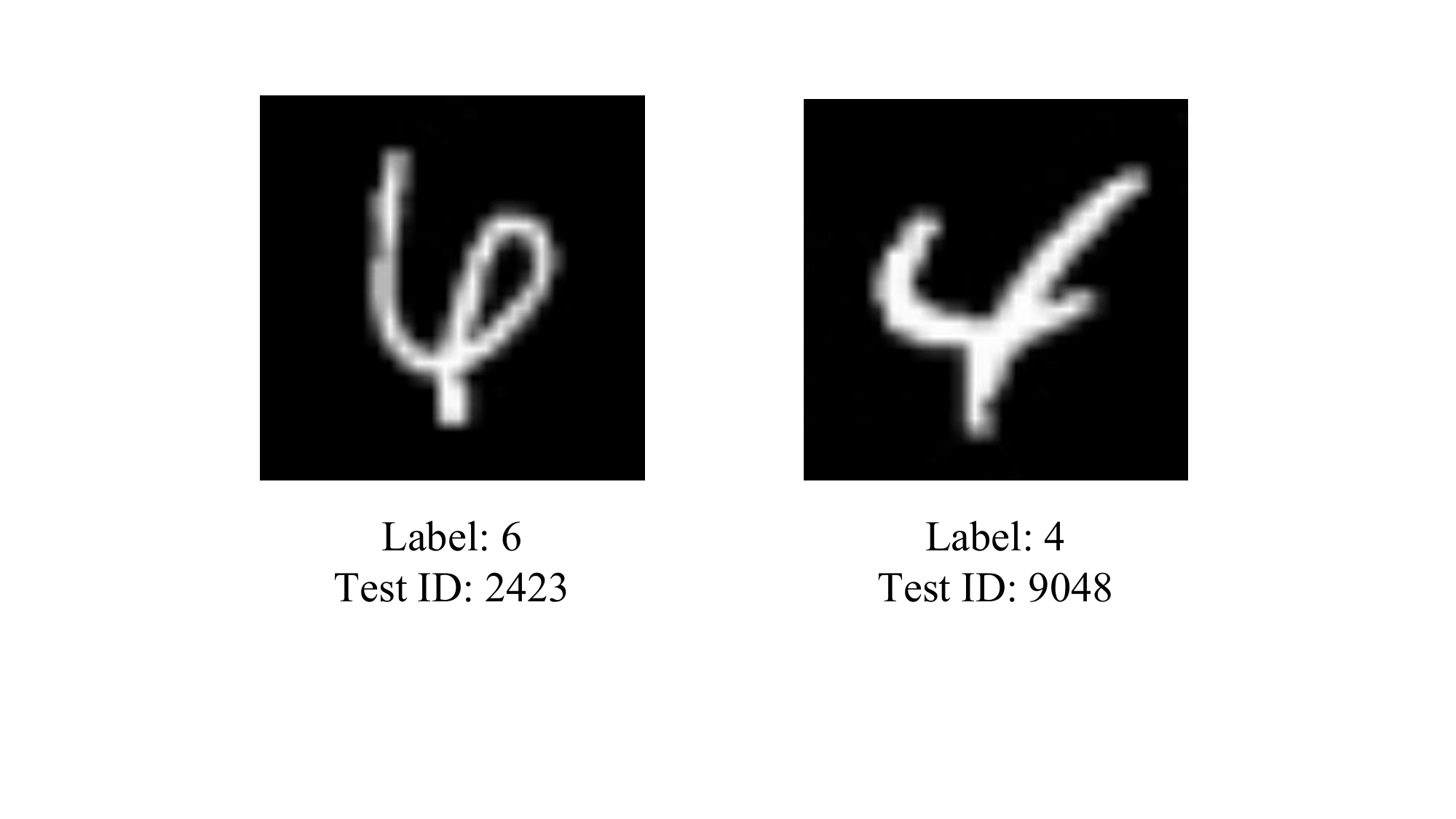}
	\caption{Ambiguous numbers in MNIST}
	\label{fig:am_mnist}
\end{figure}

For example, in the target of classification, there are many labels of dataset. In MNIST dataset, the labels are integers from $0$ to $9$ in Eq.\eqref{eq:mnist_sample}.  There is a middle stage that the input picture is not clear and it is ambiguous between $4$ and $6$ in Fig.\ref{fig:am_mnist} and corresponding event is $\{4,6\}$. Event space of MNIST is in Eq.\eqref{eq:mnist_event}. If there is $95\%$ confidence that target is in $\{4,6\}$, the mass form of confidence is in Eq.\eqref{eq:mnist_sample_mass}. 

\begin{equation}
	\begin{split}
		\Omega = \{0,1,2,3,...,9\}
	\end{split}
	\label{eq:mnist_sample}
\end{equation}

\begin{equation}
	\begin{split}
		\mathcal{E} =\{\{0\},\{1\},...,\{9\},\{0,1\},...,\Omega\}
	\end{split}
	\label{eq:mnist_event}
\end{equation}

\begin{equation}
	\begin{split}
		m(\{4,6\})=0.95
	\end{split}
	\label{eq:mnist_sample_mass}
\end{equation}

\subsection{Confirming Information Sources}\label{sec:con_inf_sou}
A source can be recorded as a set $\varpi$ of  event-mass function tuples in Eq.\eqref{eq:source} and is called Evidential Source (ES). \textbf{Information sources provide the effective data to decision system and information sources of a decision system may not be single}. Single source also has \textbf{different views} which also called \textbf{multi-modality} such as matrix, series and so on. Ensemble Learning which is an example composed of many classifiers, and each one can be seen as a source. Final decision is also single output and it is necessary to fuse different sources. 

\begin{equation}
	\begin{split}
		\varpi=\{(C,m(C))\ |\ C \in \mathcal{E} \}
	\end{split}
	\label{eq:source}
\end{equation}

\begin{figure}[htbp]
	\centering
	\includegraphics[width=0.9\linewidth]{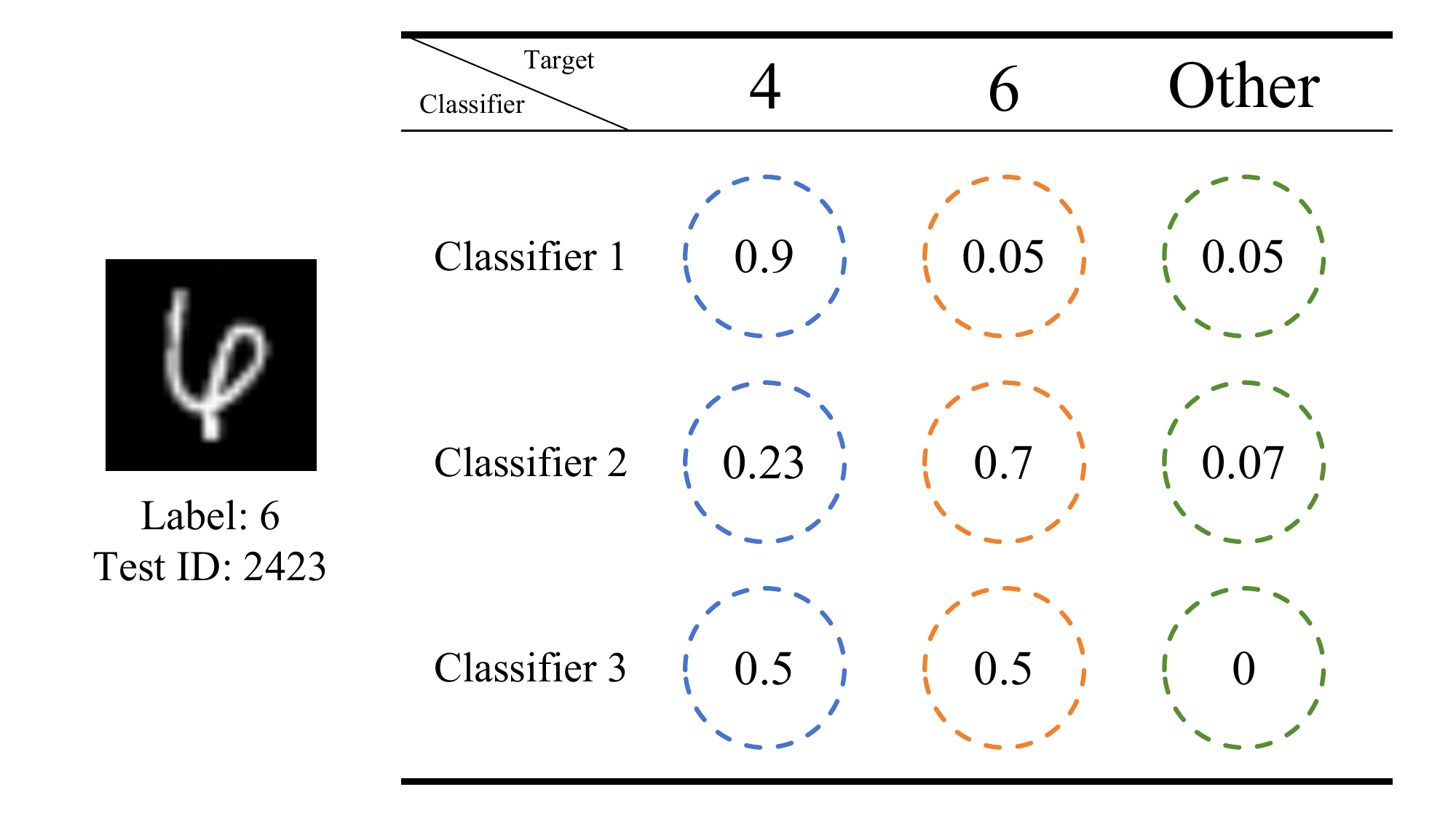}
	\caption{Three sources of input information and s}
	\label{fig:multi_source}
\end{figure}

In Fig.\ref{fig:multi_source} , there are 3 assumed classifiers about MNIST test picture 2423. The actual label  is $6$. But the preference of input picture in classifiers are different:
\begin{enumerate}
	\item \textbf{Classifier 1}: $4$ is more possible.
	\item \textbf{Classifier 2}: $6$ is more possible.
	\item \textbf{Classifier 3}: It is hard to classify $4$ and $6$.
\end{enumerate}
It is a confused problem how to determine the degree of impact of each classifiers on the final decision. That is called as \textbf{conflict} in some researches. In EPRM, conflict does not exist because \textbf{PO} treats every information source equally according to defined standards, and the handling of seemingly conflicting rules will also be defined in PO.

\subsection{Designing basic probability assignment algorithm}
BPA is a way to transform input data to mass function and different sources is able to own different BPA algorithms. Event space of BPA is the only constraint in designing of BPA algorithm that same event space ensures the same output target.

In the case of Fig.\ref{fig:multi_source}, it assumes that there are three BPA algorithms for three sources:
\begin{enumerate}
	\item \textbf{(Source 1)} Keep the top-2 probabilities and assign other probabilities equally to top-2. 
	\item \textbf{(Source 2)} Output the maximum probability. 
	\item \textbf{(Source 3)} Merge all same value probabilities.
\end{enumerate}
The generated mass function by given BPA algorithms is in Tab.\ref{tab:bpa}. In EPRM, a group of mass function generated by BPA algorithms is called as a piece of BS.

\begin{table}[htbp]
	\centering
	\caption{Mass function of three different basic probability assignment algorithms}
	\resizebox{0.95\linewidth}{!}{
	\begin{tabular}{cc}
		\hline
		Classifier & Basic Probability Assignment \\ \hline
		Classifier 1&  $m_1(\{4\})=0.925$  $m_1(\{6\})=0.075$  \\
		Classifier 2&  $m_2(\{4\})=1$  \\
		Classifier 3&  $m_3(\{4,6\})=1$   \\  \hline
	\end{tabular}}
\label{tab:bpa}
\end{table}

\subsection{Implement Pattern Operator}
PO is mentioned in Section \ref{sec:con_inf_sou} and is a binary operator which is marked as $\odot$. PO is seen as a generalized operator of DST. In DST, PO is equal to  intersection operator $\cap$ between different set. In RPS theory, PO is equal to  left/right intersection operators $\mathop{\cap}\limits^{\leftarrow} / \mathop{\cap}\limits^{\rightarrow}$ between different RPS. 

PO is not limited to intersection and any operation between events is able to be regarded as PO. Dubois and Prade proposed Disjunctive Combination Rule (DCR)  in Eq. \eqref{eq:dcr} and DCR is based on union operator $\cup$ between sets. PO is agile and can even define as repeatable set  plus operation in Eq.\eqref{eq:rset_plus} where $a_i$ is number of elements $b_j$.

\begin{equation}
		m_{\cup}(C) = \sum_{A \cup B=C} m_1(A)*m_2(B) \qquad  (\odot=\cup)
	\label{eq:dcr}
\end{equation}

\begin{equation}
	\{a_1*b_1, a_2*b_2\} \odot \{a_3*b_1, a_4*b_2\} = \{(a_1+a_3)*b_1, (a_2+a_4)*b_2\}
	\label{eq:rset_plus}
\end{equation}

\subsection{Pattern Fusion}\label{sec:pa_fu}
When PO $\odot$ is defined, fused mass function is defined in and the fused source is defined in Eq.\eqref{eq:EPRM_combination} and Eq.\eqref{eq:EPRM_combination_res}. Pattern fusion assigns the mass function of empty set to other events that ensures sum of non-empty events' mass function $\sum_{C \in \mathcal{E}} m(C)$ is equal to $1$ after calculation of Eq.\eqref{eq:EPRM_combination_res}. In other words, \textbf{EPRM does not support open world setting} because mass function of empty set $m(\emptyset)$ is always equal to $0$ and mass function of empty set does not appear in the fused source $\varpi_\odot$. Additionally, transference of empty set function is also regarded as \textbf{a type of scale normalization}. 

\begin{equation}
	m_{\odot}(C) =\left\{
	\begin{aligned}
		&\frac{\sum_{C_1 \odot C_2 ...\odot C_k=C } \prod_{i=1}^{k} m_i(C_i)}{m(\emptyset)}\quad (C \neq \emptyset)\\
		&\sum_{C_1 \odot C_2 ...\odot C_k=C } \prod_{i=1}^{k} m_i(C_i)\quad (C = \emptyset)\\
	\end{aligned}
	\right.
	\label{eq:EPRM_combination}
\end{equation}

\begin{equation}
	\begin{split}
		\varpi_\odot =
		&\{(C, \frac{\sum_{\forall C_1 \cap C_2...\cap C_k = C}\prod_{i=1}^{k}m_i(C_{i})}{m(\emptyset)}) \\
		&|\ \forall C \in \mathcal{E} \}
	\end{split}
	\label{eq:EPRM_combination_res}
\end{equation}

The algorithm of two source fusion is in Algorithm \ref{alg:two_fusion}. Because the properties of PO such as associative law and commutative law cannot be determined, the fusion of multiple information sources needs to be decomposed into the sequential fusion of two information sources.

\begin{algorithm}[htbp]
	\label{alg:two_fusion} 
	\caption{Fusion algorithm of two sources}
	\KwIn{Source 1 $\varpi_1$, Source 2 $\varpi_2$, Pattern Operator $\odot$}
	\KwOut{Fused source $\varpi$}
	\tcp{Initialize the source of fusion as an empty set}
	$\varpi=\{\}$\;
	\tcp{Initialize the mass function of empty set}
	$m_\emptyset=0$\;
	\tcp{Performing Cartesian product $\otimes$ on two sources}
	$\varpi_\otimes =\varpi_1 \otimes  \varpi_2$\;
	\tcp{Traverse each set of items - mass function tuples}
	\For{$((C_1, m_1(C_1)),(C_2,m_2(C_2))) \in \varpi_\otimes$}
	{
		\tcp{Pattern Reasoning}
		$C = C_1 \odot C_2$ \;
		\tcp{Update fused source}
		\eIf{$C \neq \emptyset$}
		{
			\eIf{$(C,m(C)) \in \varpi$ }
			{
				$\varpi \xLongleftarrow{update}(C, m(C)+m_1(C_1)*m_2(C_2))$  \;
			}{
				$\varpi \xLongleftarrow{add}(C, m_1(C_1)*m_2(C_2))$  \;
			}
		}{
			$m_\emptyset = m_\emptyset + m_1(C_1)*m_2(C_2)$
		}
		
	}
	\tcp{Normalization}
	\For{$(C, m(C)) \in \varpi$}
	{
		$(C, m(C)) \longleftarrow (C, \frac{m(C)}{m_\emptyset})$ \;
	}
	\Return $\varpi$ \;
	
\end{algorithm}

\subsection{Reasoning and Decision Making}
 Decision making is usually based on probability, so the need for fused sources is transformed into corresponding probabilities. PPT is a classical PMF transference methods in TBM that there is an assumption that  process does not utilize other information. \textbf{However, a real Decision making system is not completely without external information, and there may be subjective factors from users, so the generation of s does not require special limitations.}

Finally, the model deduces the corresponding  symbols through probability and process is also diverse. In target of image classification, a common approach is to perform one-hot encoding on labels, then optimize them using cross entropy, and finally use the label with the highest probability as the  result. \textbf{To summarize the DMO $DM(\cdot)$ , it can be considered that it consists of two parts: one is the fused source $\varpi$, and the other is external information or hyperparameters $\omega$ of user preferences in Eq.\eqref{eq:EPRM_DM}.} When the hyperparameters $\omega$ are an empty set $\emptyset$, it can be considered that the current system only considers the input raw data without any preferences, and \textbf{PPT is an example of no preferences}.

\begin{equation}
	\begin{split}
		Decision =  DM(\varpi, \omega). 
	\end{split}
	\label{eq:EPRM_DM}
\end{equation}

\section{Random Graph Set}\label{sec:RGS}

\subsection{Relation Expression Of Graph}
\begin{figure}[htbp]
	\centering
	\includegraphics[width=0.98\linewidth]{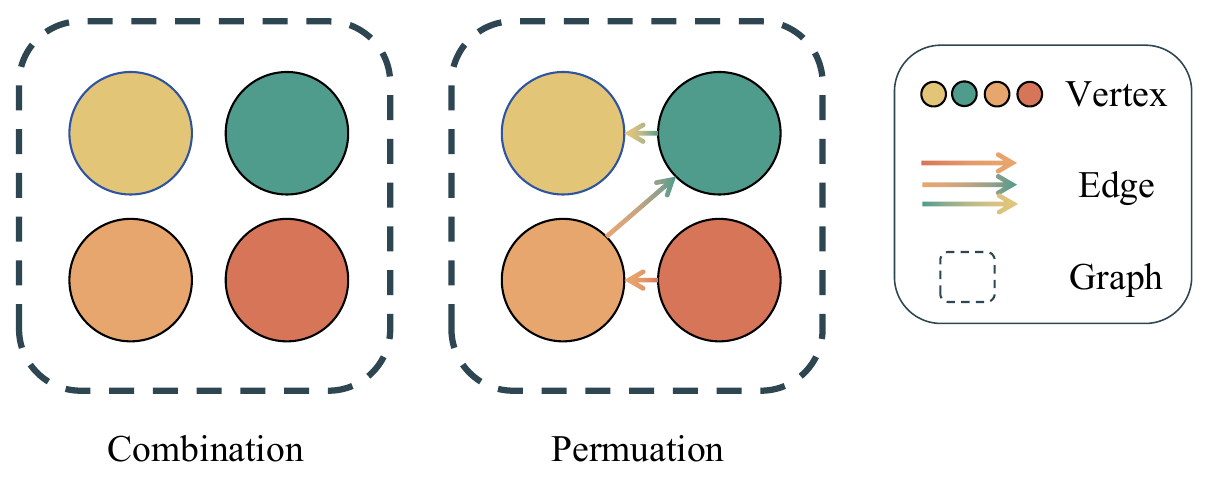}
	\caption{Combination and permutation in form of graph}
	\label{fig:graph}
\end{figure}

\begin{figure}[htbp]
	\centering
	\subfloat[Cycling Event]
	{
		\label{fig:sub_cyc}\includegraphics[width=0.47\linewidth]{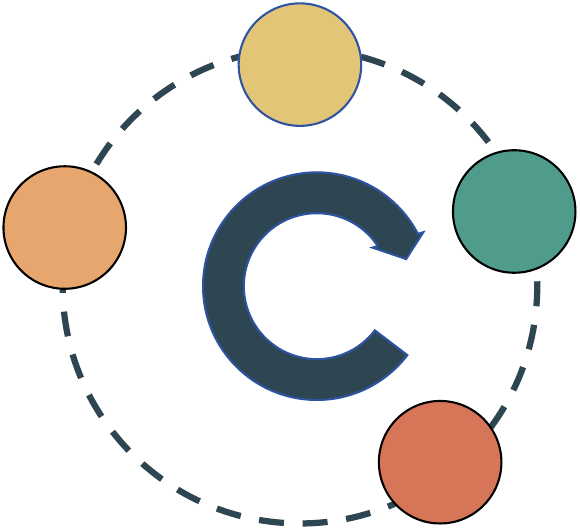}
	}
	\subfloat[Parallel Event] 
	{
		\label{fig:sub_para}\includegraphics[width=0.4\linewidth]{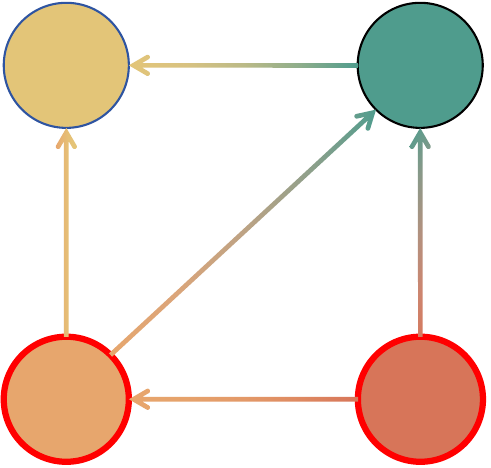}
		
	}
	\caption{Advantages of graph expression}
	\label{fig:graph_ad}
\end{figure}

Event space expresses the relations between samples in sample space. Combination relation is considered by DST, and permutation relation is considered by RPS theory. However, simple combination and permutation are not complete for modeling real-world data. Therefore, Random Graph Set (RGS) was proposed and modeled using a graph approach.  A graph is a structure that can directly represent relationships and is compatible with arrangement and combination relationships In Fig.\ref{fig:graph}. \textbf{When there are no edges in the graph, the graph represents a combination relationship. In addition, when the graph is a single directed chain, the graph represents order information.} For some special relationships, permutations and combinations cannot be directly used for representation, such as cycles and parallel operations in Fig.\ref{fig:graph_ad}. The advantage of graphs is that they are compatible with more relationship types, making the modeling process more efficient.

\subsection{Definition}
Sample space  $\Omega$ of RGS is as same as DST and RPS that composes of atomic existed samples $V_i$ in Eq.\eqref{eq:rgs_ss}. Event space $\mathcal{E}$ of RGS is a set composed of graph $G$ where $V$ is a set of vertexes and $E$ is a set of edges in $G$ in Eq.\eqref{eq:rgs_es}. ES of RGS is in Eq.\eqref{eq:rgs_bs}.

\begin{equation}
	\Omega  = \{V_1,V_2,V_3,...,V_n\}
	\label{eq:rgs_ss}
\end{equation}

\begin{equation}
	\mathcal{E}  = \{G(V, E)\ | \ V \in \Omega \}
	\label{eq:rgs_es}
\end{equation}

\begin{equation}
	\mathcal{\varpi}  = \{(g_i,m(g_i)) \ | \ g_i \in \mathcal{E} \}
	\label{eq:rgs_bs}
\end{equation}

\begin{figure*}[htbp]
	\centering
	\includegraphics[width=0.7\linewidth]{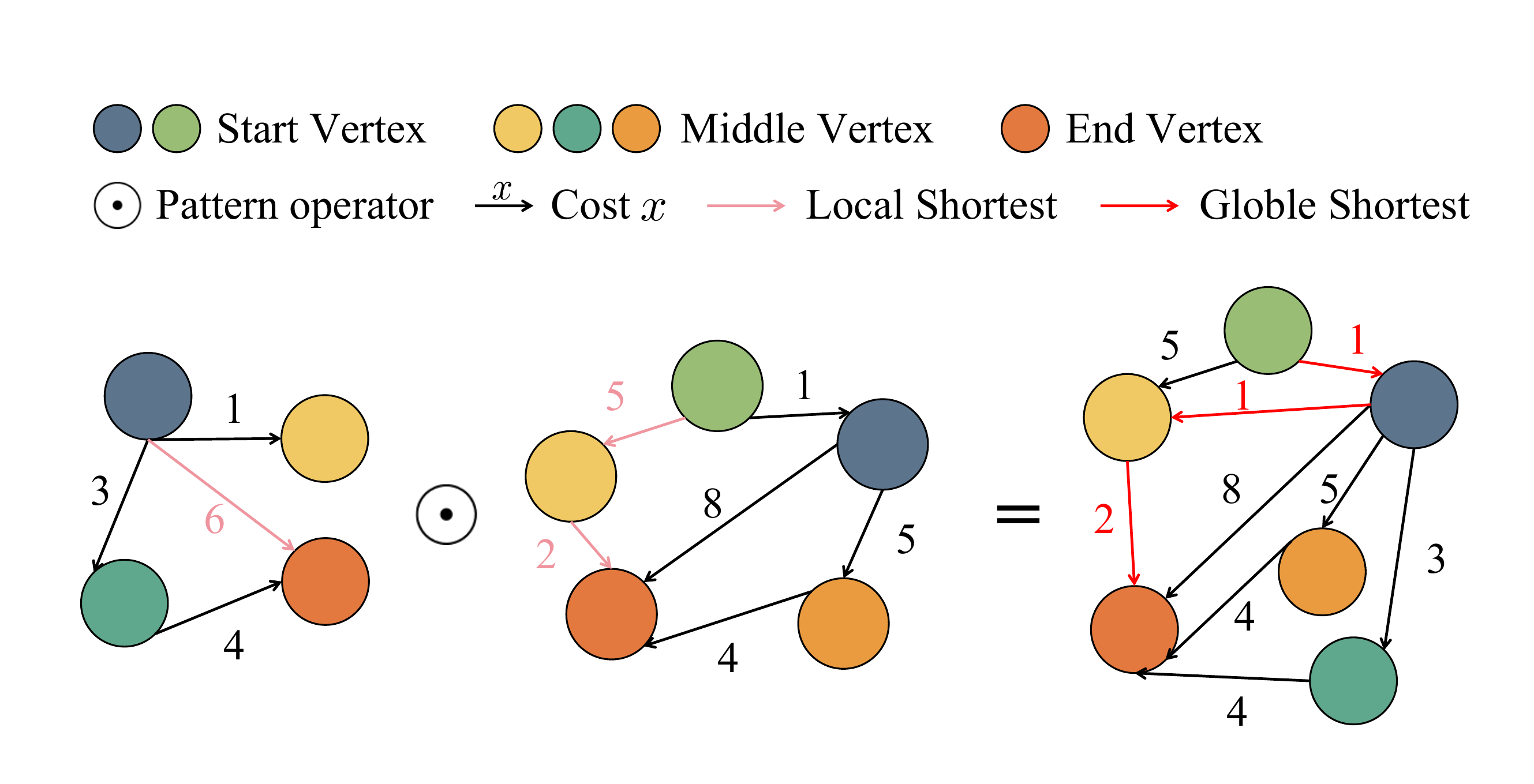}
	\caption{Example of two events fusion by PO: the minimum cost }
	\label{fig:po_rgs}
\end{figure*}

\subsection{Pattern Operator Of Random Graph Set}
Through EPRM, graph operations can be easily transferred to PO the reasoning process of EPRM. There are many operations for graphs, such as union and intersection. For complex problems, if the optimization objective is to minimize the cost, the PO can be defined as \textbf{first merging two graphs and then finding the shortest path among them}. At this point, the  PO $\odot$ is the concatenation of graph merging and shortest path algorithm in Fig.\ref{fig:po_rgs}.

\section{Aircraft Velocity Recognition}\label{sec:air}
The flight velocity of an aircraft is influenced by many factors. In order to effectively identify the position and velocity of the aircraft, multiple sensors can be used for position detection. The experiment is divided into two parts:
\begin{enumerate}
	\item \textbf{(Trajectories Simulation)} By generating aircraft trajectories with similar average velocitys that are difficult to distinguish. Due to the fluctuation of aircraft velocity, it is difficult to distinguish the theoretical velocity of the aircraft. Therefore, \textbf{the target task is to sort the aircraft velocity}. 
	\item \textbf{(Sensor Reasoning)} Assuming that the sensor locates the aircraft within its range at intervals. Assuming that sensors locate aircraft within a range at a time interval, the collected data can only distinguish which aircraft it comes from. Then use EPRM to sort the order of the aircraft's flight velocity.
\end{enumerate}

\subsection{Symbol Explanation And Basic Settings}
There are symbols in Trajectories Simulation and Sensor Reasoning part which shows in Tab.\ref{tab:air_sym}.

\begin{table}[htbp]
	\centering
	\caption{Symbols in aircraft velocity recognition}
	\resizebox{0.9\linewidth}{!}{
		\setlength{\tabcolsep}{12pt}
		\begin{tabular}{cc}
			\hline
			Variable	&  Symbol  \\ \hline
			Starting coordinates of aircraft & $(x_a,y_a)$   \\
			Direction vector of flight velocity & $(d_x, d_y)$   \\
			Current flight velocity of the aircraft & $\vec{v}$ \\
			Variance of velocity & $\sigma^2$ \\
			Period of velocity change & $T$ \\
			Current time & $t$ \\
			Detection radius of the sensor & $r$   \\ 
			Detection time interval of the sensor & $\Delta t$ \\
			Normal distribution & $N(\mu,\sigma^2)$ \\ 
			Uniform distribution & $U(a,b)$ \\ 
			Cartesian products &$\times$ \\ \hline
	\end{tabular}}
	\label{tab:air_sym}
\end{table}

\begin{figure}[htbp]
	\centering
	\includegraphics[width=0.98\linewidth]{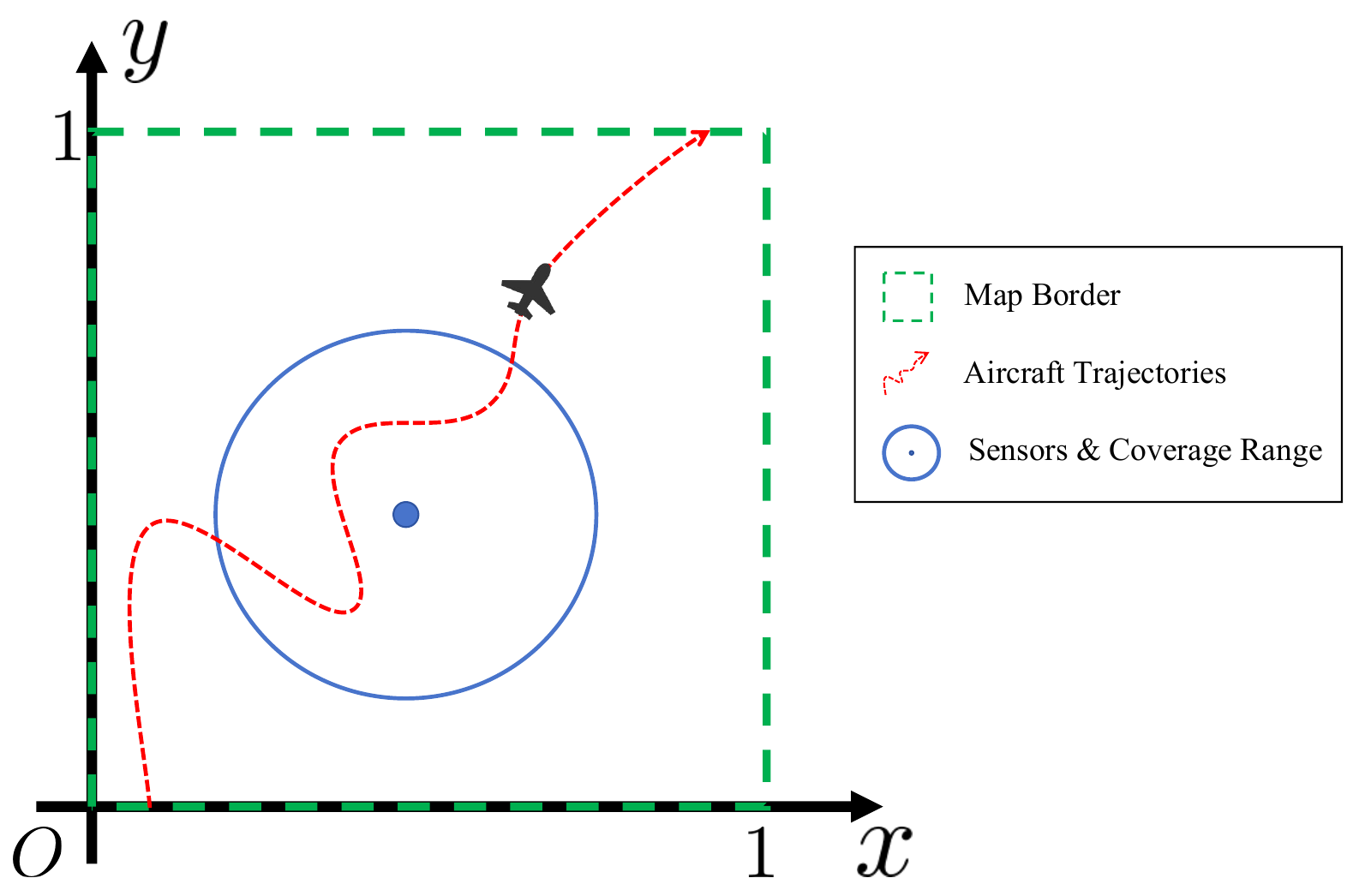}
	\caption{Map settings for simulations}
	\label{fig:map}
\end{figure}

In order to standardize the simulation of aircraft trajectories, the global map of the simulation will be limited to a square with a side length of $1$ in Fig.\ref{fig:map}. Time and velocity are also normalized.

Because the ideal velocities of different aircraft are similar, there is significant disturbance in the actual velocities that \textbf{sensors only need to infer the velocity ranking of an aircraft}. Sensors may not be able to detect certain aircraft due to their coverage range, or their data may conflict. Therefore, \textbf{effective reasoning only needs to ensure that the order is consistent with the theoretical velocity ranking, which can be considered as a part of the theoretical ranking.} If the reasoning result is single aircraft or pure aircraft without ranks, it can be considered invalid as no velocity related information has been provided.

\subsection{Trajectories Simulation}
The method of trajectory simulation is stochastic process. The velocity $\vec{v}$ of an aircraft is a vector, which is orthogonally decomposed into two directions: x axis $\vec{v_x}$ and y axis $\vec{v_y}$ in Eq.\eqref{eq:velocity_decompose}. The components of velocity are stochastic processes that are time-dependent in Eq.\eqref{eq:velocity_com_sim}. By accumulating the velocity $\vec{v}$, the position $(x(t),y(t))$ of the aircraft can be simulated in Eq.\eqref{eq:pos_sim}.

\begin{equation}
	\vec{v}(t) = \vec{v_x}(t) + \vec{v_y}(t) 	
	\label{eq:velocity_decompose}
\end{equation}

\begin{equation}
	\left\{
	\begin{aligned}
		& |\vec{v_x}(t)| = d_x * v_{xr}(t) \quad  v_{xr}(t) \sim N(|\vec{V}|, \sigma^2*\frac{|sin(2\pi t)|}{T} ) \\
		& |\vec{v_y}(t)| = d_y* v_{yr}(t) \quad  v_{yr}(t) \sim N(|\vec{V}|, \sigma^2*\frac{|sin(2\pi t)|}{T} ) \\
	\end{aligned}
	\right.
	\label{eq:velocity_com_sim}
\end{equation}

\begin{equation}
	\left\{
	\begin{aligned}
		&  x(t) = \sum_{i=0}^{\frac{t}{\Delta t}}  |\vec{v_x}(i*\Delta t)|*\Delta t + x_a\\
		&  y(t) = \sum_{i=0}^{\frac{t}{\Delta t}}  |\vec{v_y}(i*\Delta t)|*\Delta t+ y_a\\
	\end{aligned}
	\right.
	\label{eq:pos_sim}
\end{equation}

The detection range of a sensor is a circle and the radius is $r$. In order to ensure that the coverage range of the sensor is completely within the map, the simulation of the sensor coordinates $(x,y)$ is in Eq.\eqref{eq:pos_sensor}.

\begin{equation}
	\left\{
	\begin{aligned}
		&  x \sim U(r,1-r)\\
		&  y \sim U(r,1-r)\\
	\end{aligned}
	\right.
	\label{eq:pos_sensor}
\end{equation}

\subsection{Sensor Reasoning}
The sensors detect the temporal coordinates of each aircraft. So, each sensor can calculate the average velocity of each aircraft at each time interval in Eq.\eqref{eq:sensor_velocity}.

\begin{equation}
	\vec{v}(t) = \frac{\sqrt{(x(t+\Delta t) - x(t))^2 + (y(t+\Delta t) - y(t))^2}}{\Delta t}
	\label{eq:sensor_velocity}
\end{equation}

It is generally believed that the average value can effectively reflect the theoretical velocity of the aircraft, but due to the small number of collected points and similar aircraft velocitys, the following two decision options will be compared: \textbf{MVD} and \textbf{CRD}.

\subsubsection{Mean Value Decision}
Mean Value Decision is a direct method that uses the average velocity $\bar{|v|}$ of each flying object detected by the sensor as the basis for velocity ranking. Each sensor will independently give the aircraft's velocity ranking.

\subsubsection{Conflict Resolution Decision}
Multiple sensors may conflict in the velocity ordering of the aircraft due to random disturbances. Therefore, CRD method is proposed. CRD is an implementation of EPRM. The following will introduce step by step how to implement CRD:

\begin{enumerate}
	\item \textbf{(Confirming Sample Space And Event Space)} There are $n$ aircraft in the sample space, where $n$ is the number of aircraft in the case. The event space $\mathcal{E}$ is a RGS with aircraft as nodes $V$. The graph in RGS is an unweighted directed graph, in which the directed edge $(a \rightarrow b)$ indicates that the theoretical velocity of $a$ is \textbf{smaller} than that of $b$.
	\item \textbf{(Confirming Information Sources)} The information sources are different instantaneous velocity data collected by $m$ sensors.
	\item \textbf{(Designing basic probability assignment algorithm)} The generation of BPA and the calculation of probability are not exactly the same. Similar to the conversion of frequency and probability, it needs to count the aircraft velocity first in Algorithm \ref{alg:sp_cou}. 
		
	\begin{algorithm}[htbp]
		\label{alg:sp_cou} 
		\caption{velocity aircraft interval count}
		\KwIn{The velocity $v_i$ of each aircraft detected by the sensor}
		\KwOut{Interval count $num$}
		\tcp{Merge all velocitys}
		$V_m=\{v_{m,i}\ | \ v_{m,i} \in \bigcup_{j=1}^{n} v_j \}$\;
		\tcp{Initialize interval count}
		$num = \{\} $ \;
		\tcp{Traverse all velocitys}
		\For{$v_{m,i} \in V_m$}
		{
			\tcp{Initialize the belonging interval}
			$belong = \{\} $ \;
			\tcc{Traverse the interval of each aircraft: Determine which aircraft may have generated the current velocity}
			\For{Each aircraft $j$}
			{
				\tcp{Determine whether it is within the current aircraft interval}
				\If{$MIN(v_j)\leq v_{m,i} \leq MAX(v_j)$}
				{
					$belong \xLongleftarrow{add} j$
				}
			}
			\tcp{Skip if the velocity belongs to all velocity intervals}
			\If{$belong=\{1,2,3,...,n\}$}
			{
				Continue \;
			}
			\tcp{Update the count of all subsets}
			\For{$belong_s \subseteq belong$}
			{
				\eIf{$(belong_s, num(belong_s) \in num$)}
				{
					$num \xLongleftarrow{update} (belong_s, num(belong_s)+1)$\;
				}
				{
					$num \xLongleftarrow{add} (belong_s, 1)$\;
				}
			}
		
		}
		\Return{$num$}
		\end{algorithm}
		
		The core idea of Algorithm \ref{alg:sp_cou} is to \textbf{count the aircraft velocity intervals that each velocity may belong to}. If it \textbf{belongs to the velocity intervals of all aircraft}, it means that this data does \textbf{not technically provide any information that can distinguish the aircraft in the range.} 
		
		If a velocity data belongs to multiple aircraft intervals, it can be considered that it is difficult to distinguish which velocity it belongs to. Therefore, the indistinguishable categories were merged, and the merged mean was recalculated as a reference for velocity ranking. At the same time, the count $num$ is converted into a mass function through normalization in Algorithm \ref{alg:con_sou}. Algorithm \ref{alg:con_sou} is equivalent to completing the mapping $\varrho$ of a velocity attribute category to a graph $G(V,E)$, and at the same time completing the conversion of frequency $f(i)$ to mass function $m(G(V,E))$ in Eq.\eqref{eq:c2g}, Eq.\eqref{eq:fre} and Eq.\eqref{eq:mass_g}. Mapping $\varrho$ is bijective. So all the sensors produced corresponding BPA and became ES.
		
		\begin{algorithm}[htbp]
			\label{alg:con_sou} 
			\caption{Count converted to evidence source}
			\KwIn{velocity $v_i$ of each aircraft detected by the sensor, Interval count $num$}
			\KwOut{Evidence source $\varpi$}
			\tcp{Initialize the evidence source}
			$\varpi = \{\}$ \;
			\tcp{Traverse velocity count}
			\For{$(c, num(c)) \in num$}
			{
				\tcp{Initialize empty graph as event , the nodes are all detected aircraft}
				$g = G(V=\{1,2,...,n\}, E=\{\})$ \;
				\tcp{Calculate the average of distinguishable aircraft}
				\eIf{$|c|=1$}
				{
					$A = Avg(\{v_1,v_2,...,v_n\}) $ \;
				}{
					$A =Avg(\{v_1,v_2,..., v_n\} -\{v_i\ | \ i \in c\}+c)$
				}
				\tcc{
					Sort aircraft by mean velocity \\
					$e.g.$ \\
					$|c|=1$: $order = \{\{1\},\{3\},\{2\}\}$ \\
					$|c|=2>1$: $order = \{\{2,3\},\{1\}\}$}
				$order = sort(A)$ \;
				\tcp{Traversal order}
				\For{$i \in \{1,2,...,|order|-1\}$}
				{
					\tcc{
						Construct edges using Cartesian products \\
						$e.g.$ \\
						$\{2,3\} \times \{1\} = \{(2,1),(3,1)\}$
						}
					$E_s = order(i) \times order(i+1)$ \;
					\tcp{Add the constructed edge to the event}
					\For{$e \in $}
					{
						$E \xLongleftarrow{add} e$
					}
					\tcp{Calculate mass function}
					$m = num(c)/SUM(num)$ \;
					\tcp{Update evidence sources}
					\eIf{$(g,m(g)) \in \varpi$ }
					{
						$\varpi \xLongleftarrow{update}(g, m(g)+m)$  \;
					}{
						$\varpi \xLongleftarrow{add}(g, m)$  \;
					}
				}
			}
			\Return{$\varpi$}
		\end{algorithm}
		
		\begin{equation}
			\varrho: c \rightarrow G(V,E)
			\label{eq:c2g}
		\end{equation}
		
		\begin{equation}
			f(i) = \frac{num(i)}{\sum_{j \in \varpi} num(j)}
			\label{eq:fre}
		\end{equation}
		
		\begin{equation}
			m(G(V,E)) = \sum_{\varrho(c) \subseteq k, \varrho(k)=G(V,E)} f(c)
			\label{eq:mass_g}
		\end{equation}
	\item \textbf{(Pattern Fusion)} In the Section \ref{sec:pa_fu}, pattern fusion Algorithm \ref{alg:two_fusion} is proposed and in velocity sorting applications only PO $\odot$ need to be implemented. Since the velocity patterns generated by different sensors may conflict, the principle of graph fusion is to sort the fusion results according to the \textbf{frequency of occurrence} in Algorithm \ref{alg:po_air}. $\odot$ will select the edge with the most detection times from a set of directed edges in opposite directions to avoid conflicts. The opposite direction of the directed edges means that the relationship between the detected velocitys of the two aircraft is not constant but varied. There may be loops in the fused graph $g$. Because of the directed graph $g$, and in the absence of bidirectional edges, loops have their own flow direction. The loop removal algorithm $rm\_loop$  deletes the directed edge in the loop that starts from the node that reaches the furthest path for the first time. Similar to this, the short path deletion algorithm $rm\_short\_path$ will only retain the longest one when there are multiple paths between any two points in the graph, ensuring the sorting information during the detection process to the greatest extent.
	
	\begin{algorithm}[htbp]
		\label{alg:po_air} 
		\caption{Implementation of PO $\odot$ in aircraft velocity sorting}
		\KwIn{Graphs $g_i$ from ES $\varpi_i$: $\{g_1(V_1,E_1),g_2(V_2, E_2),...,g_n(V_n,E_n)\}$}
		\KwOut{Fused graph $g$}
		\tcp{Initialize the set of edges count}
		$E_c = \{\}$, $E_c$ is a repeatable set\;
		\For{Edge $e=(start, end)$ in $\bigcup_{i=1}^{n} E_i$}
		{
			\eIf{$\bar{e}=(end, start) \notin E_c$}
			{
				$E_c \xLongleftarrow{add} 1*e$ \;
			}{
				$E_c \xLongleftarrow{remove} 1*\bar{e}$ \;
			}
		}
		\tcp{Convert repeatable set to normal set}
		$E = \{ e\ | \ k*e \in E_c\}$ \;
		\tcp{Get the node that constructs the edge}
		$V =\{v\ | \ (v,k) \in E\ \&\ (k,v) \in E\}$ \; 
		\tcp{Constructing fused graph}
		$g =G(V,E)$ \;
		\tcp{Remove edges of loop}
		$g = rm\_loop(g)$ \;
		\tcp{Keep the longest path between any two nodes and delete other possible paths}
		$g = rm\_short\_path(g)$ \;
		\Return{$g$}
	\end {algorithm}
	
	The above-mentioned operations of constructing and modifying the fusion graph $g$ all deal with conflicts between detected velocity information, and avoid conflicts based on the principles of frequency of occurrence first and length of sorted information first. It is worth noting that the detected velocitys are indeed conflicting, because the actual velocity will be affected and not necessarily consistent with the theoretical velocity ranking.
	
	\item \textbf{(Reasoning and Decision Making)} The way of reasoning is to make decisions based on the largest mass function. Since there may be multiple maximum mass functions, only the conflict-free parts of these graphs will be retained. Finally, the conflict-removed graph $g$ is separated, and the decision result is the path from all starting points to the end points of $g$. The corresponding algorithm is Algorithm \ref{alg:dm}.
	
		\begin{algorithm}[htbp]
		\label{alg:dm} 
		\caption{Implementation of decision making $DM$ in aircraft velocity sorting}
		\KwIn{Fused ES $\varpi$}
		\KwOut{Decision $D$}
		\tcp{Initialize decision}
		$D = \{\}$ \;
		\tcp{List the group of graphs with the largest mass}
		$G =\mathop{argmax} \limits_{(g, m(g)) \in \varpi }(m(g))$ \;
		\tcp{Exclude conflicting rules}
		\eIf{$|G|>1$}
		{
			$g = g_1 \odot g_2.... \odot g_n \quad (g_i \in G)$ \;
		}{
			$g = g_i  \quad (g_i \in G)$ \;
		}
		\tcp{Get the nodes in the graph}
		$n = node(g)$ \;
		\tcp{The starting point of filtering, $ID(\cdot)$ refers to the in-degree of the node.}
		$start = \{\ n_i \ | \  ID(n_i)=0 \ \& \  n_i \in n\}$  \;
		\tcp{The ending point of filtering, $OD(\cdot)$ refers to the out-degree of the node.}
		$end = \{\ n_i \ | \  OD(n_i)=0 \ \& \  n_i \in n\}$ \;
		\tcp{Separate decision order, $\times$ refers to Cartesian products}
		\For{$(s,e) \in start \times end $}
		{
			\tcp{Get the path from the starting point to the end point, $Path(\cdot,\cdot)$ refers to the path between two nodes}
			$path = Path(s,e)$ \;
			$D \xLongleftarrow{add} path$ \;
		}
		\Return{D}
		\end {algorithm}
\end{enumerate}

The calculation process of CRD is complicated because there are many relationship features represented by the graph and there are many conflicts that exist at the same time. At the same time, since the decision making goal is order, using graphs in reasoning can represent more order-related features. 

\begin{figure*}[htbp]
	\centering
	\subfloat
	{
		\includegraphics[width=0.245\linewidth]{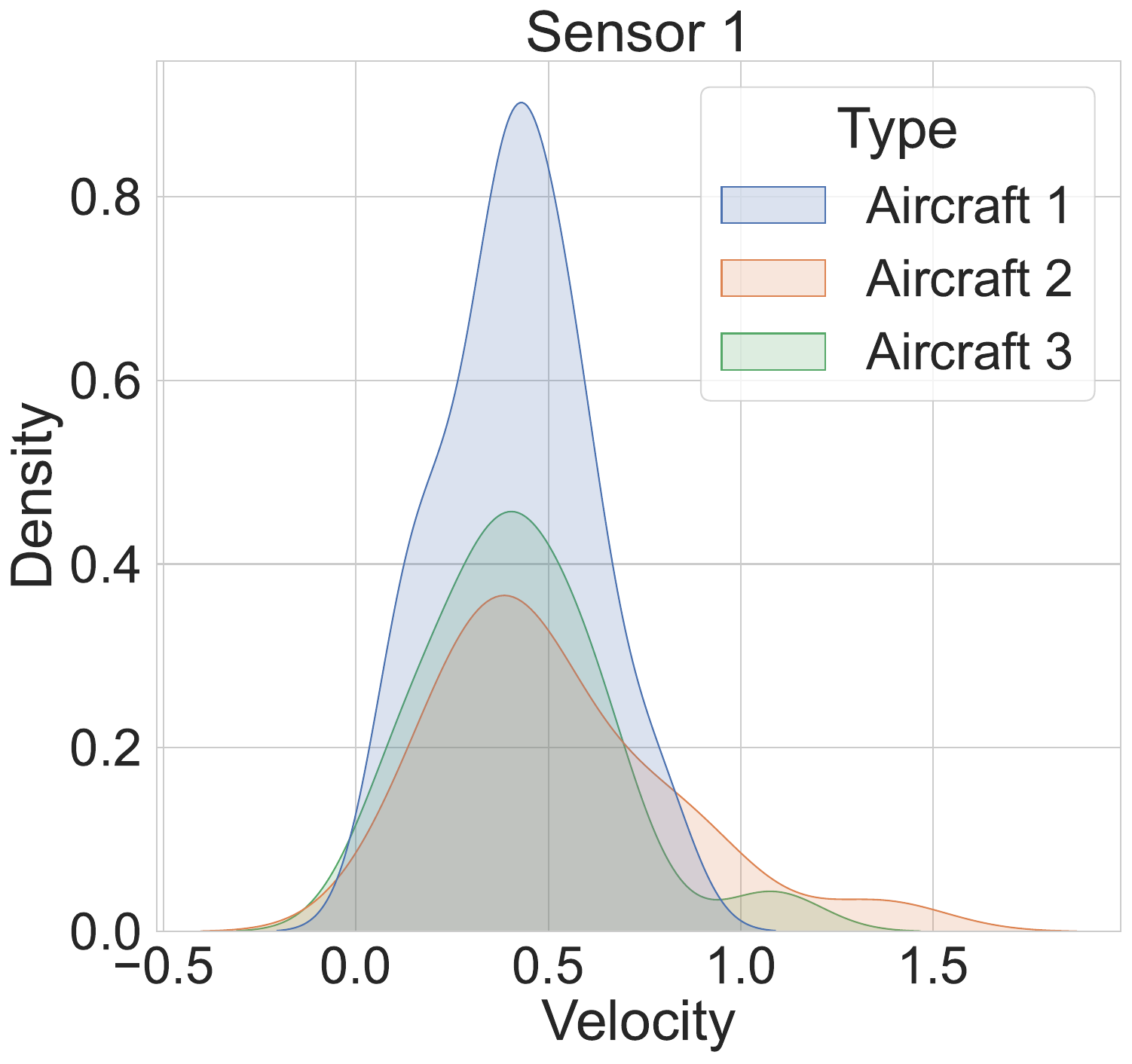}
	}
	\subfloat
	{
		\includegraphics[width=0.245\linewidth]{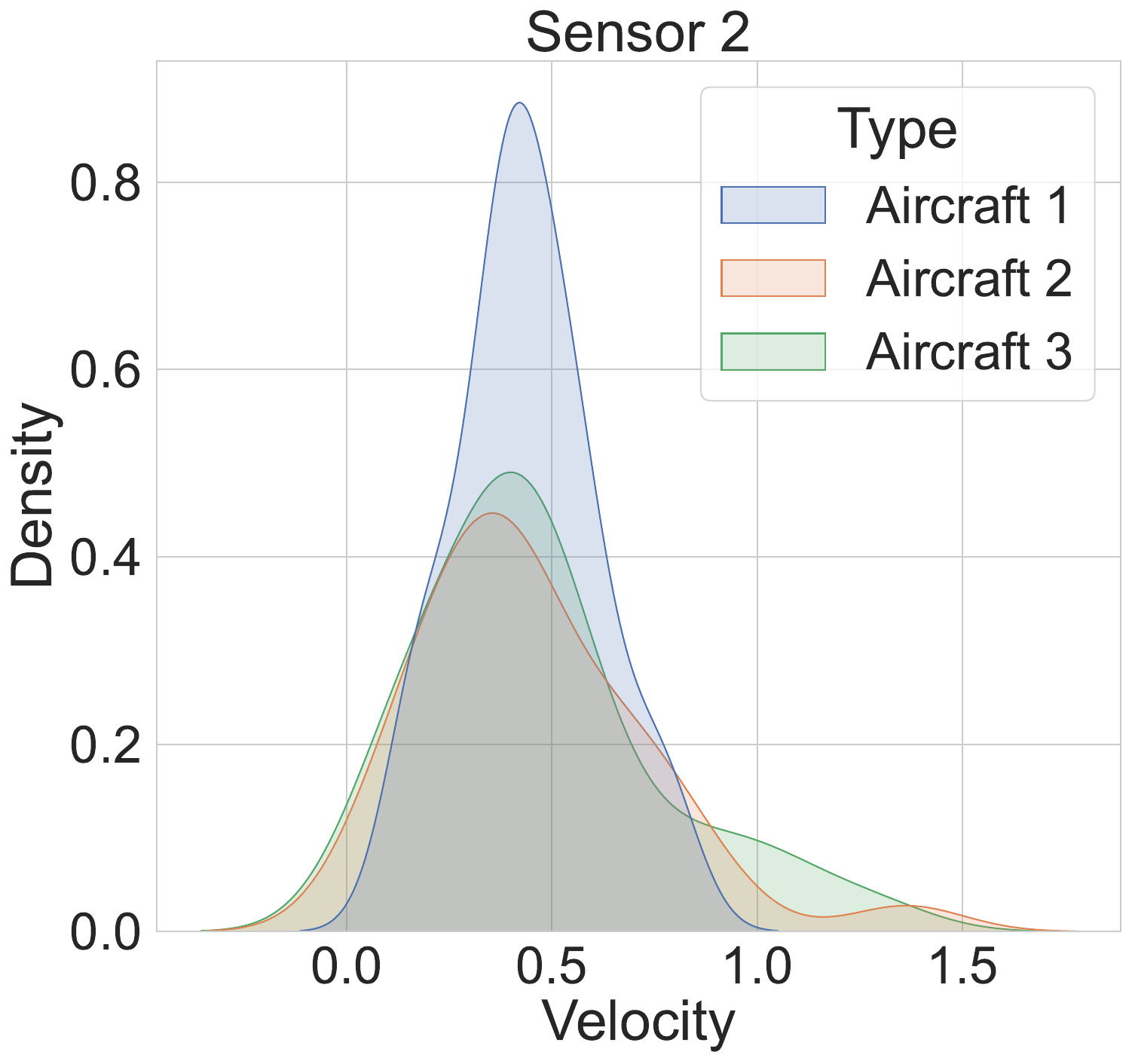}
	}
	\subfloat
	{
		\includegraphics[width=0.245\linewidth]{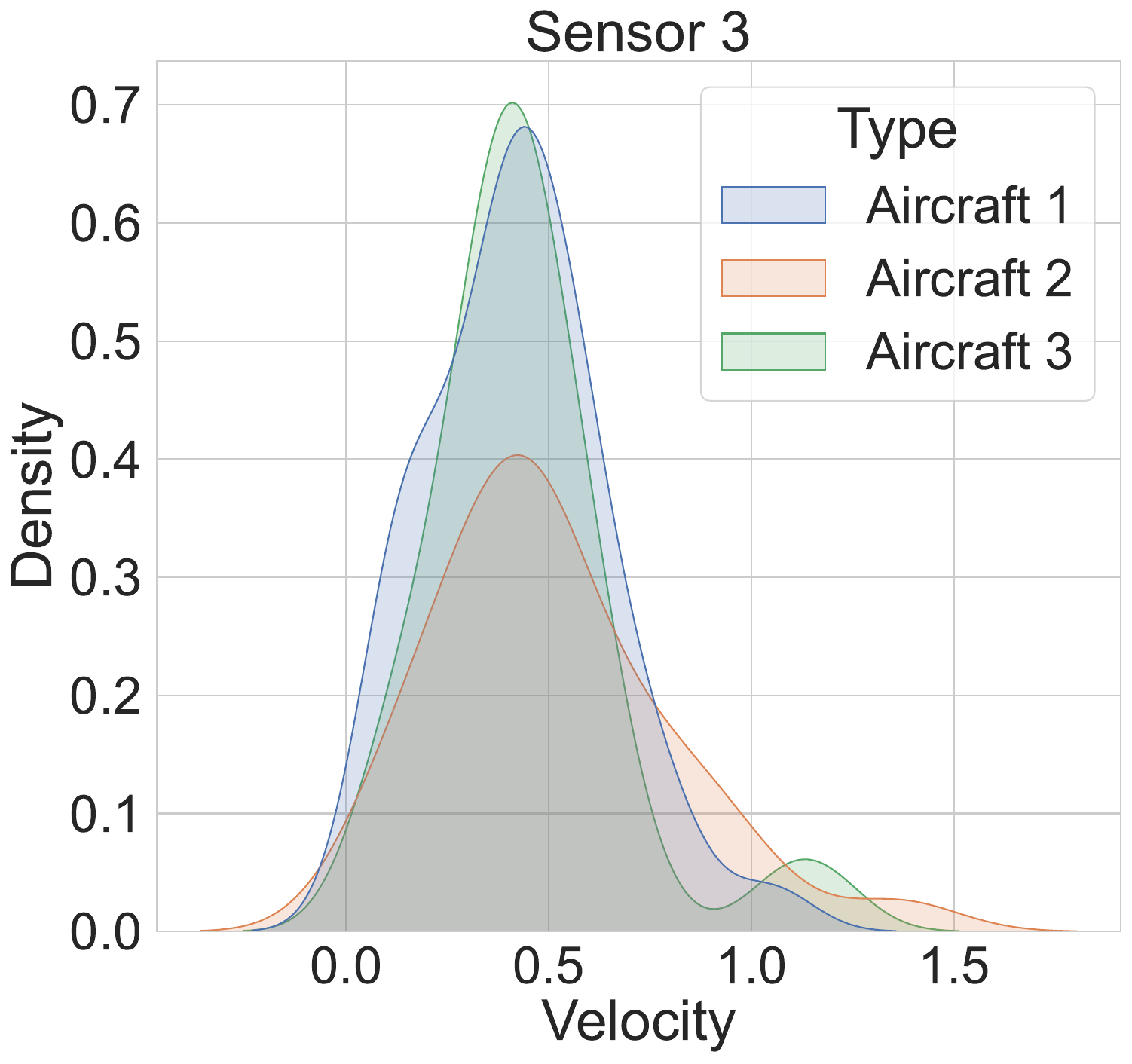}
	}
	\subfloat
	{
		\includegraphics[width=0.245\linewidth]{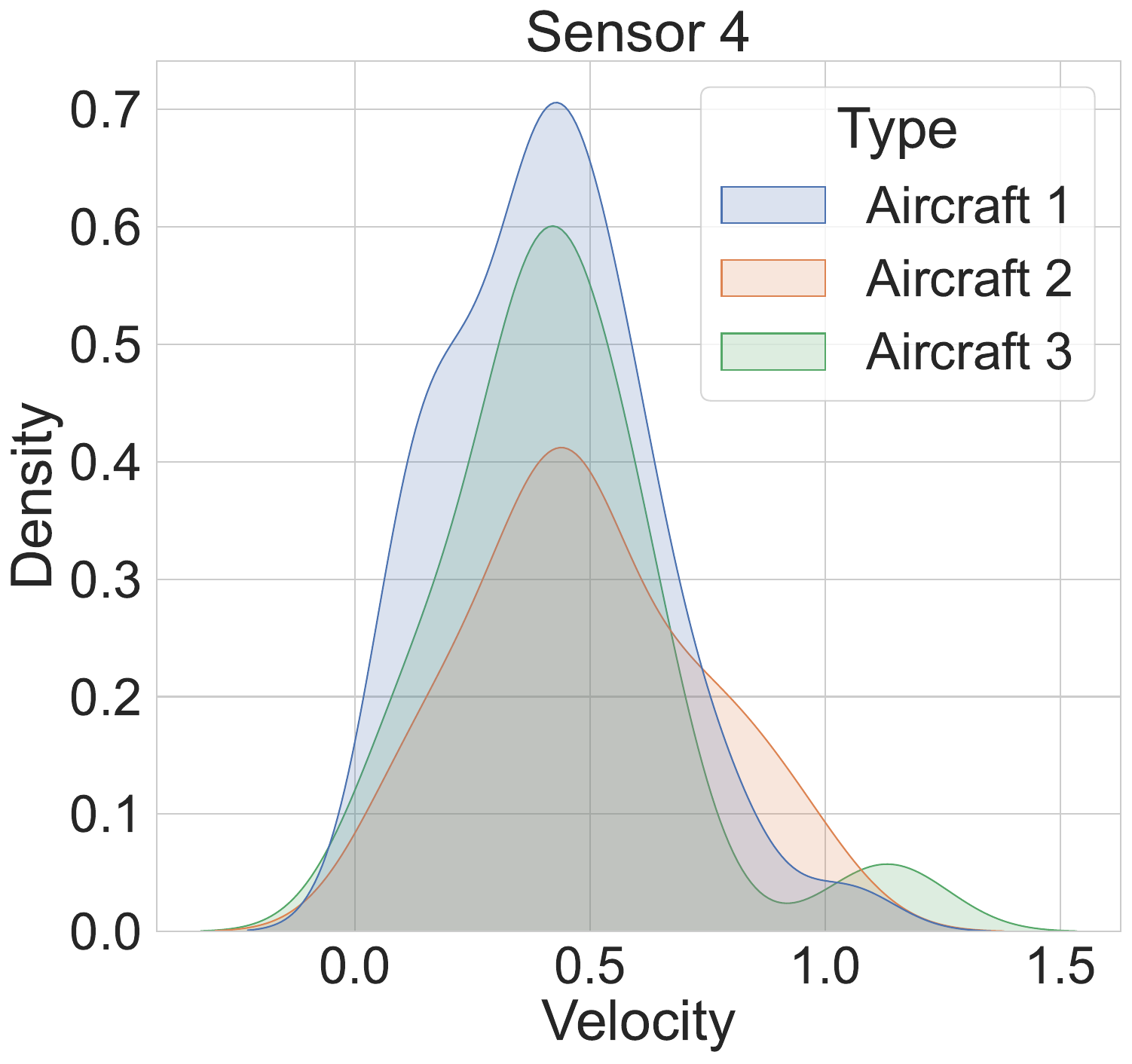}
	}
	\\
	\subfloat
	{
		\includegraphics[width=0.245\linewidth]{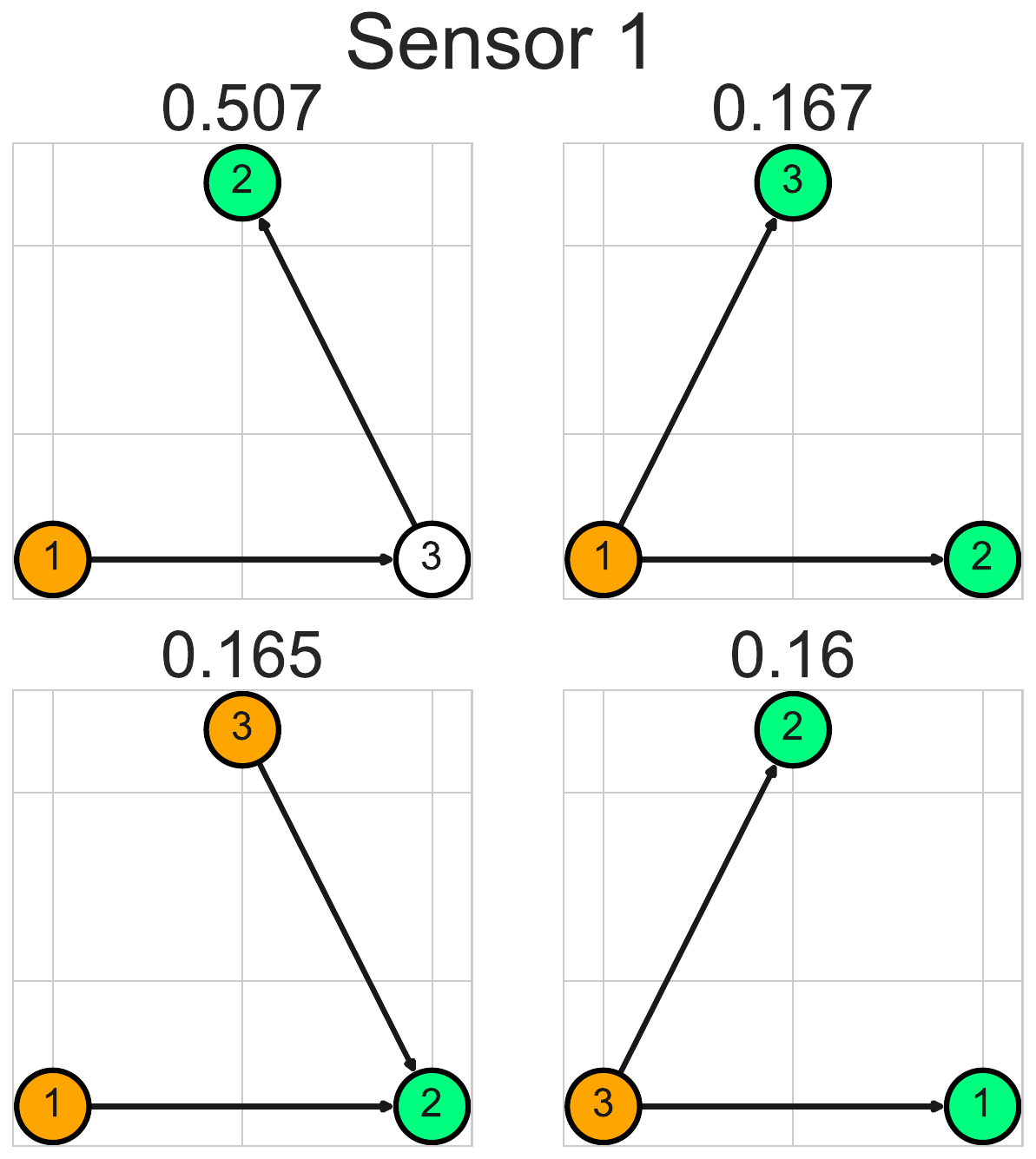}
	}
	\subfloat
	{
		\includegraphics[width=0.245\linewidth]{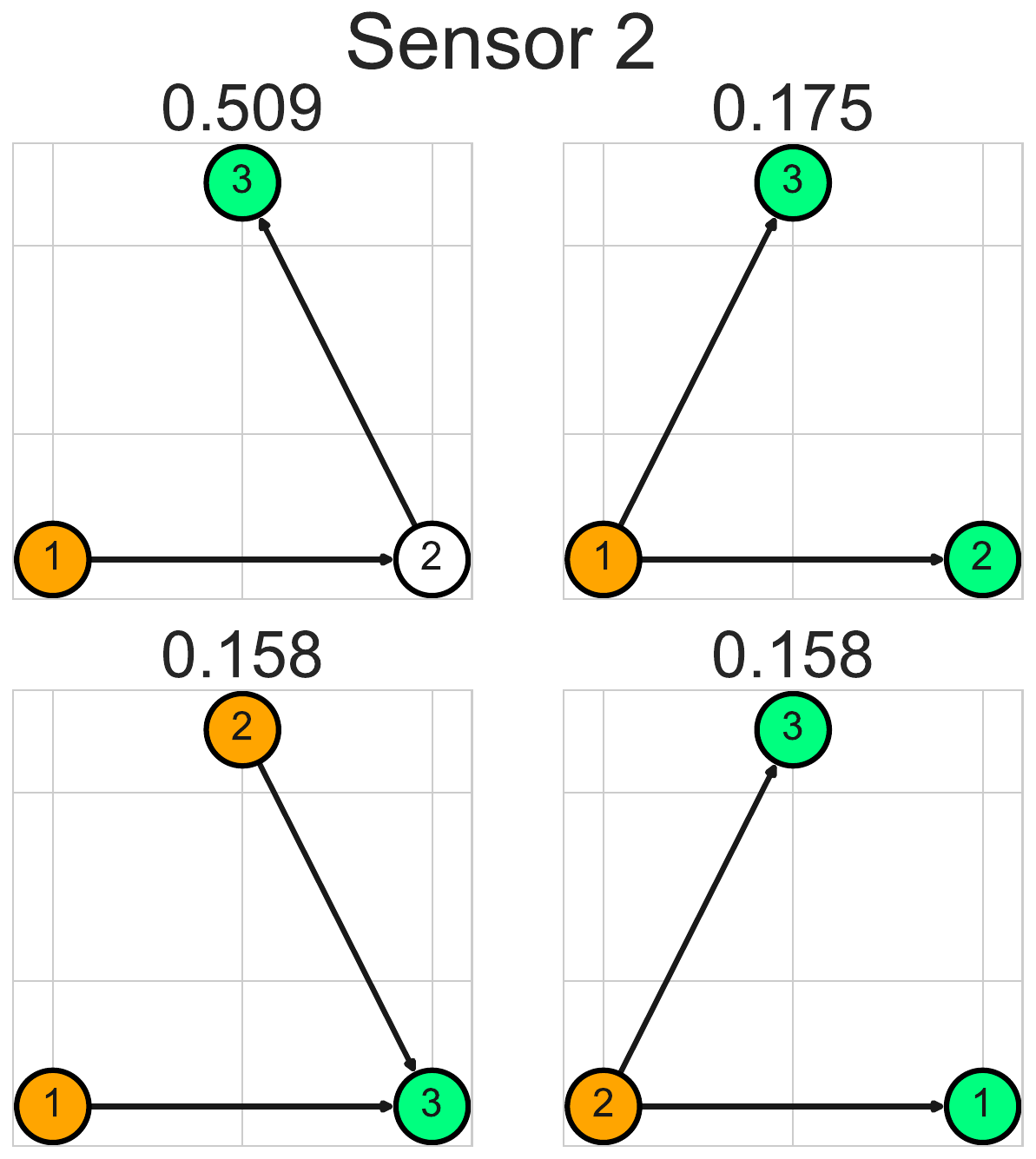}
	}
	\subfloat
	{
		\includegraphics[width=0.245\linewidth]{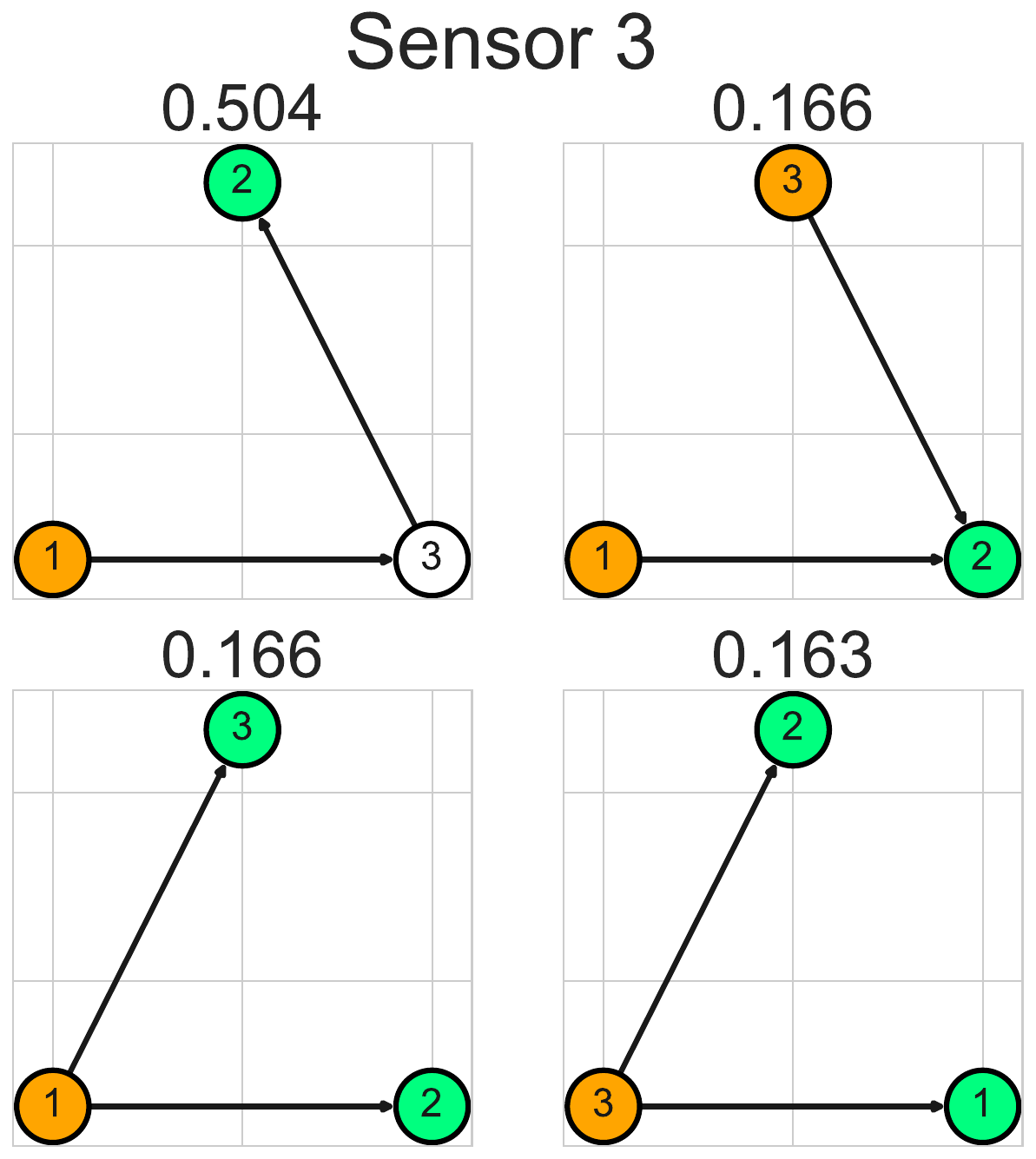}
	}
	\subfloat
	{
		\includegraphics[width=0.245\linewidth]{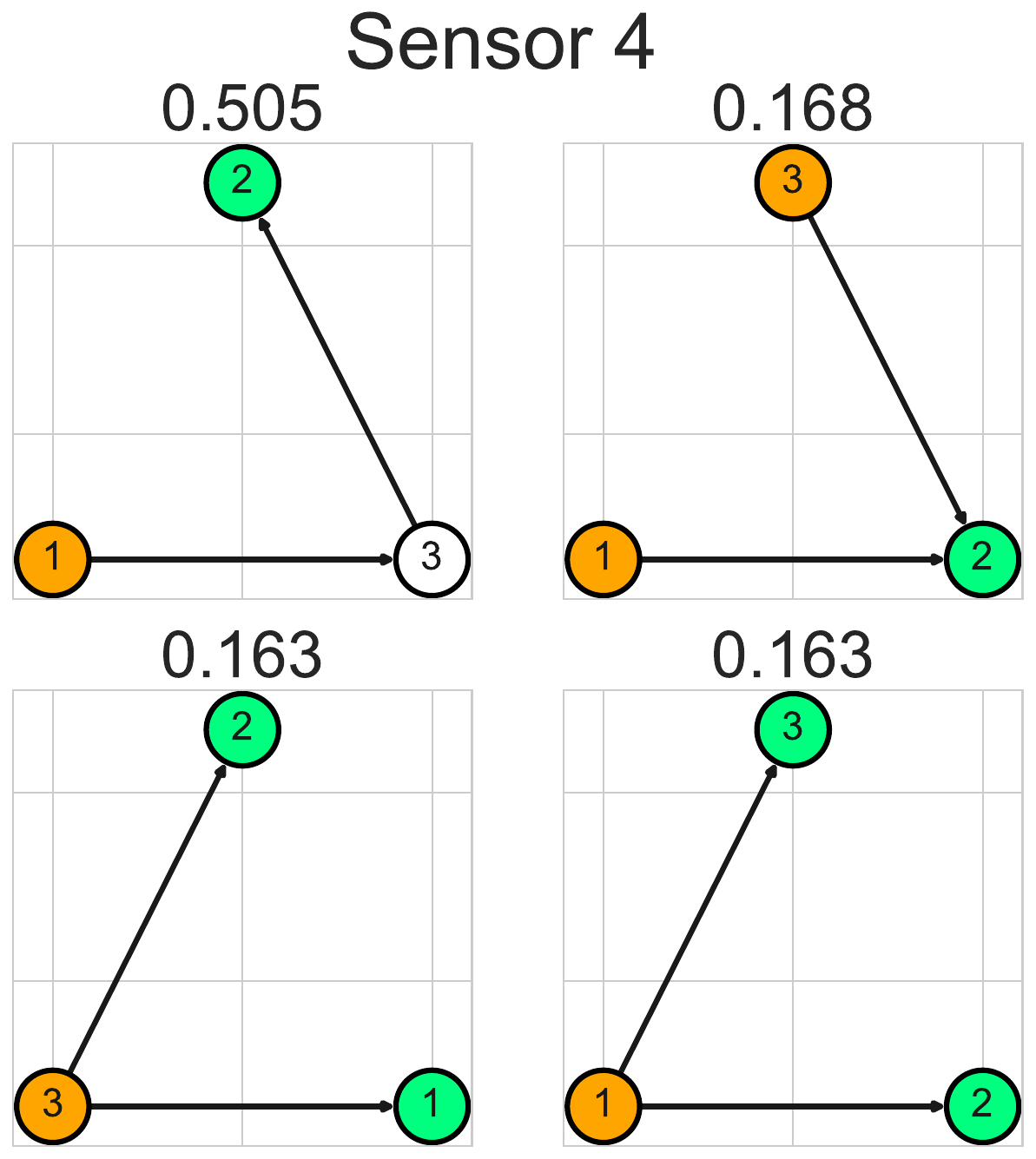}
	}
	\caption{Sensor detection data and corresponding evidence sources}
	\label{fig:crd_dis}
\end{figure*}

\begin{table*}[htbp]
	\centering
	\caption{Average velocity sorted by sensor }
	\resizebox{0.95\linewidth}{!}{
		\setlength{\tabcolsep}{12pt}
		\begin{tabular}{ccccc}
			\hline
			Sensor	  &  Order  &  Aircraft 1 Mean Velocity & Aircraft 2 Mean Velocity & Aircraft 3 Mean Velocity \\ \hline
			Sensor 1	&  \textcolor{blue}{$1<3<2$} &  $0.418$ & $0.423$ & $0.516$ \\ 
			Sensor 2	&  \textcolor{red}{$1<2<3$} &  $0.44$ & $0.453$ & $0.470$ \\ 
			Sensor 3	&  \textcolor{blue}{$1<3<2$} &   $0.423$ & $0.436$ & $0.504$ \\ 
			Sensor 4	&  \textcolor{blue}{$1<3<2$} &  $0.411$  & $0.433$ & $0.491$ \\  \hline
	\end{tabular}}
	\label{tab:sensor_mean}
\end{table*}

\begin{figure}[htbp]
	\centering
	\includegraphics[width=0.9\linewidth]{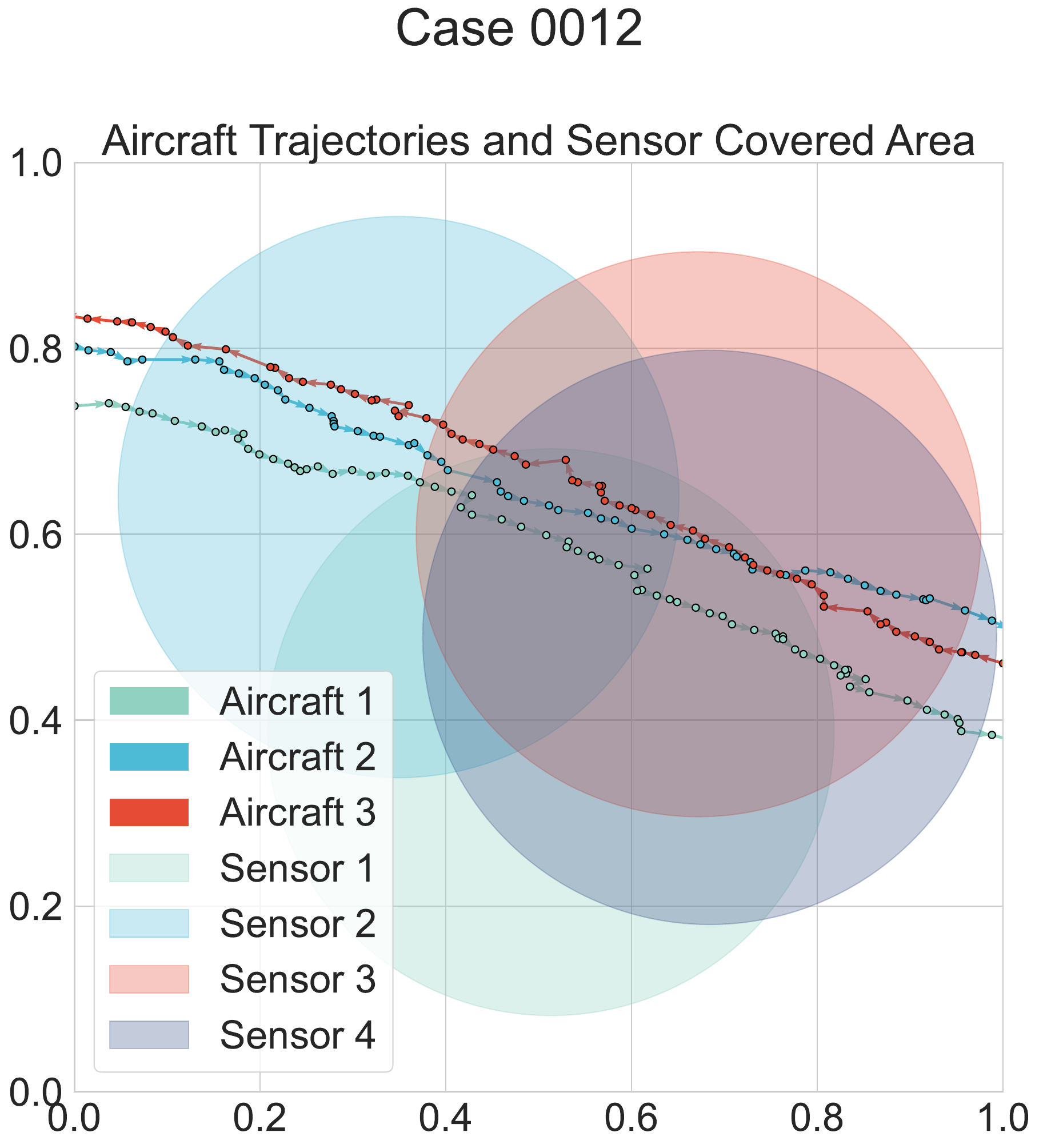}
	\caption{Map of Case 12}
	\label{fig:sim_map}
\end{figure}

\subsection{Simulated Data}
According to the mentioned simulation method, the experiment simulated 10,000 aircraft velocity detection cases. The relevant data is open sourced (\url{https://doi.org/10.21227/pcvh-0j22},  \url{https://www.heywhale.com/mw/dataset/65cf4af7f231de650fbacc0f}) \cite{pcvh-0j22-24}.  In order to explain the decision making process more specifically, Case 12 will be visualized as an example.

First of all, Fig.\ref{fig:sim_map} is the map of Case 12. The trajectories of the aircraft are drawn by several vectors connected end to end, indicating the flight direction and the flight distance of the sensor per unit time. The circle represents the detection range of the sensor.  Multiple sensors detected the aircraft's trajectory data. Tab.\ref{tab:sensor_mean} records the average velocity data of aircraft recorded by different sensors and the corresponding sorting. It can be seen that Sensor 2 and other sensors have reached conflicting conclusions. The mean velocities of aircraft measured by the three sensors that obtained the same conclusion are also different.

\begin{figure}[htbp]
	\centering
	\includegraphics[width=0.6\linewidth]{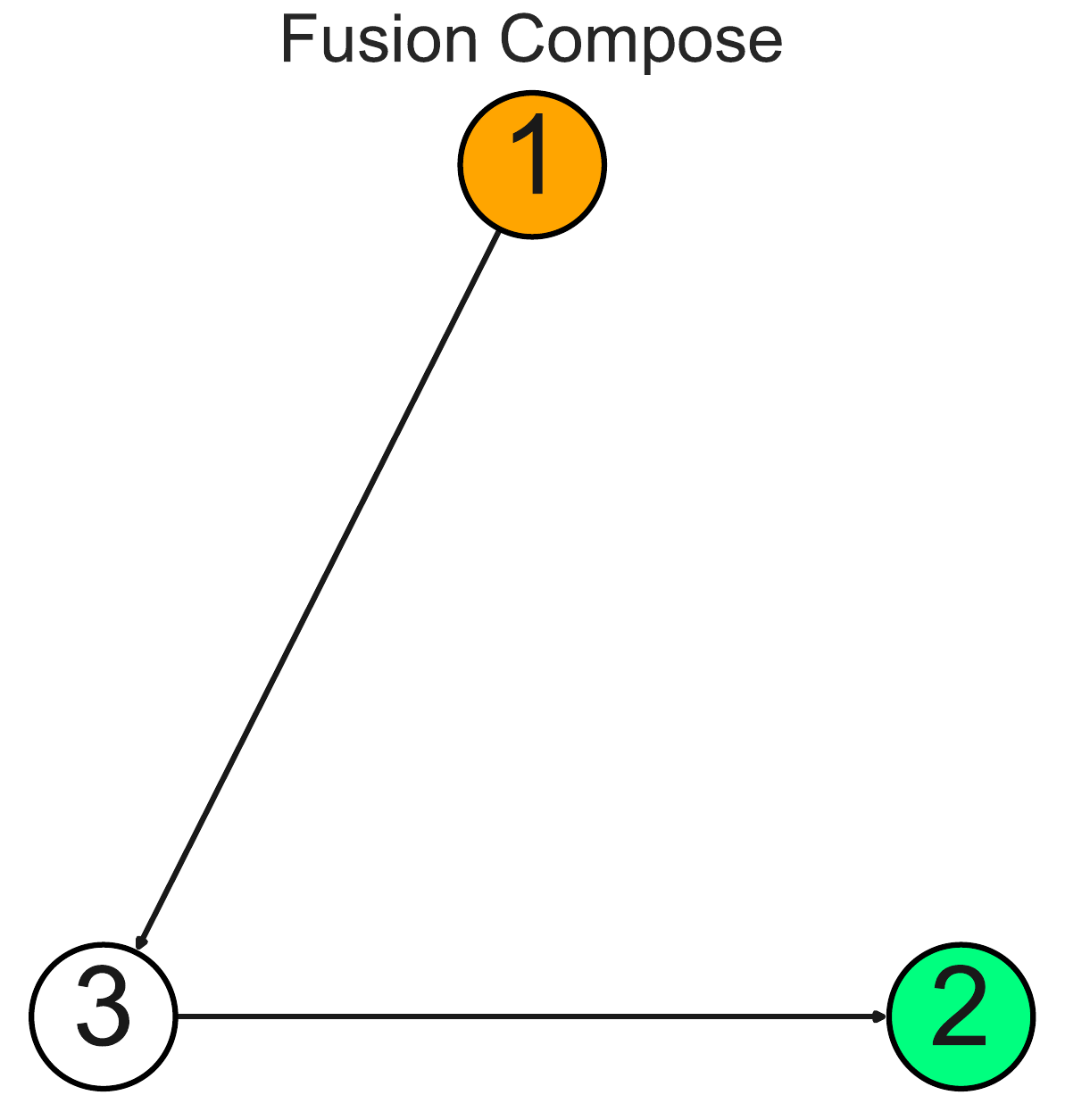}
	\caption{Decision of Case 12 with CRD}
	\label{fig:crd_dec}
\end{figure}

Compared with MVD, CRD analyzes the velocity distribution in more detail in Fig.\ref{fig:crd_dis}. According to the definition of CRD, the possible attributes of different velocities were first counted. The first row is the velocity distribution measured by each sensor. It can be seen from the density that the amount of velocity data collected by the sensor is uneven, especially for Aircraft 2. This is also the reason for conflicts, because less velocity will lead to larger errors. According to Algorithm \ref{alg:sp_cou} and Algorithm \ref{alg:con_sou}, the ES obtained by each sensor is the second row that the orange node is the starting point and the green one is the ending point. The maximum preference of ES obtained by the sensor is consistent with the average velocity, so CRD can be regarded as an optimized version of MVD. Finally, the decision result $1<3<2$ obtained through ESRM fusion is shown in Fig.\ref{fig:crd_dec}.

Case 12 shows the details in aircraft velocity sorting, including map generation, velocity measurement and analysis and decision making that is just a sample. The statistical data of the 10,000 simulated samples will be analyzed below.

\subsection{Statistical Analysis}
The results of 10,000 random simulations can effectively reflect the effect of the proposed CRD. First, how to classify the results needs to be explained. Tab.\ref{tab:state_ex} lists several statuses of the results. It is worth noting that if multiple decisions are all False, the final state will be False. If both False and True exist in multiple decisions, it is Conflict.

\begin{table*}[htbp]
	\centering
	\caption{Result State Definition}
	\resizebox{0.95\linewidth}{!}{
		\setlength{\tabcolsep}{12pt}
		\begin{tabular}{cc}
			\hline
			Result	&  Explanation  \\ \hline
			True &   The relative position of any two nodes is consistent with the theoretical ordering. \\ 
			False &  There are two nodes whose relative positions are inconsistent with the theoretical ordering.\\
			Conflict & The same edge in different ES has opposite directions.\\ 
			Invalid &  No information provided. \\ \hline
	\end{tabular}}
	\label{tab:state_ex}
\end{table*}

\begin{figure}[htbp]
	\centering
	\includegraphics[width=0.98\linewidth]{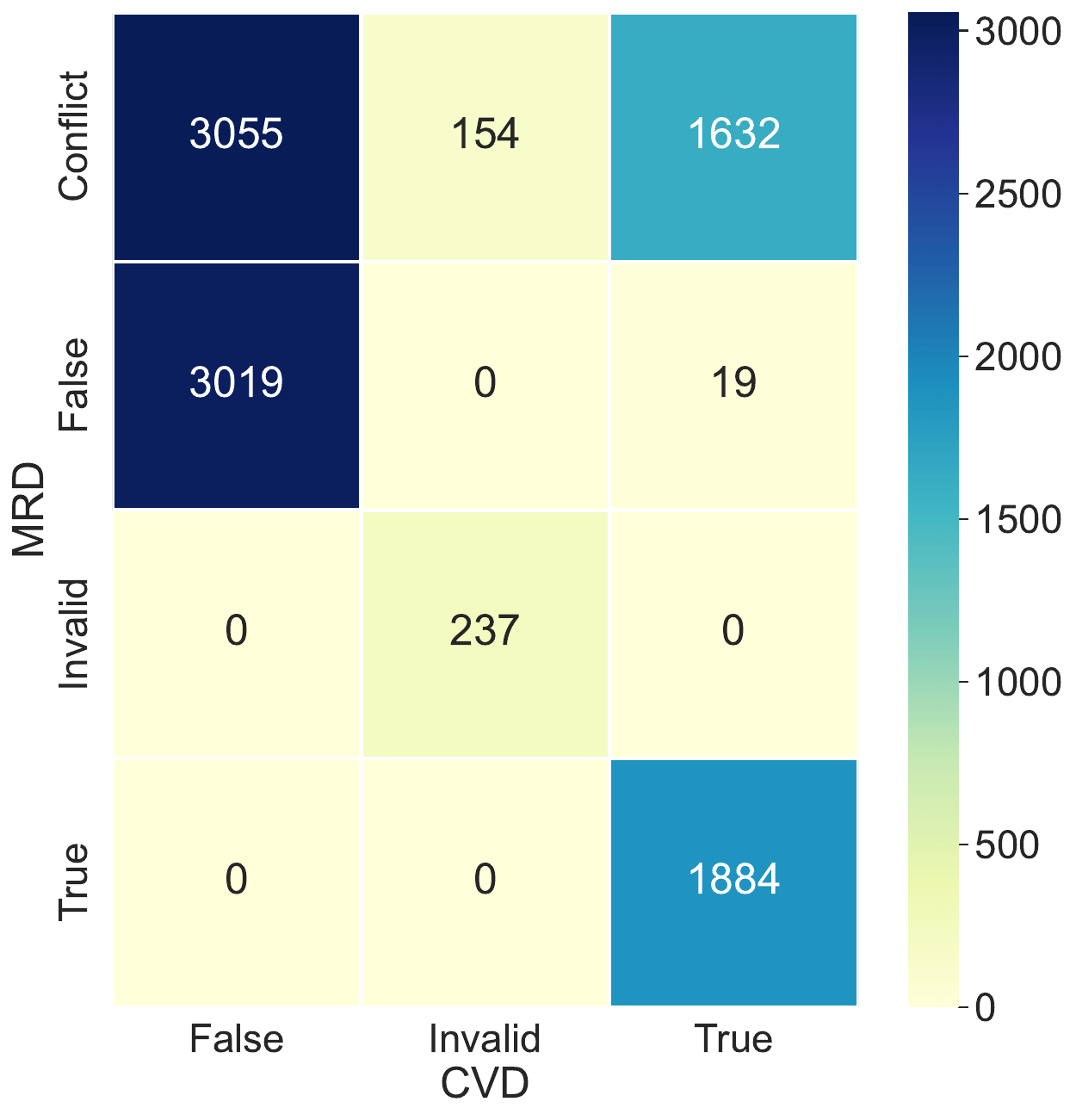}
	\caption{Heatmap of decision making results for MVD and CRD}
	\label{fig:heatmap}
\end{figure}

The decisions of 10,000 simulation samples were classified, and the results were analyzed with heat maps of the two decision making methods in Fig.\ref{fig:heatmap} that the numbers marked are counts. In terms of quantity, $78.79\%$ of the cases showed False or Conflict in the MVD. It shows that MVD cannot make good decisions on the order of velocity. \textbf{Because the simulated velocity values of different aircraft are similar and fluctuate greatly, CRD may not be able to obtain better results. The main reason is that there is a large deviation in the data collection itself, and the sequential decision making is also wrong. It is not the reason for the MVD and CRD algorithms.} The experiment needs to focus on the decision of how much CRD is optimized compared to MVD.

The cases where MVD is wrong or conflicting and CRD is wrong account for 60.74\%. Most of the missing 18.05\% cases from $78.79\%$ mistakes of MVD  are cases where CRD decided MVD as Conflict and the decision was correct after optimization. \textbf{In addition, when the decision result of MVD is True, CRD does not produce False and Invalid, indicating that CRD is completely an optimization algorithm of MVD, or that the lower limit of the effect of CRD is the same as MVD.} It can be shown on the heat map that the optimization effect of CRD is significant with blue fill. Other areas filled with yellow have little reference significance for the experiment.

\section{Conclusion}\label{sec:conclusion}
In order to solve the problem of previous evidence theory being unable to set preferences in reasoning and decision making, EPRM was proposed to better fit the target tasks and set subjective preferences. PO and DMO are where preferences are worked in the model. At the same time, in order to characterize more complex relationships between samples, RGS was proposed, which expanded the pure combination and permuation relationships in evidence theory and RPS.

The experimental section simulated $10,000$ aircraft velocity ranking cases. The statistical analysis results showed that the MVD showed a high error or conflict rate (78.79\%). The proposed CRD algorithm optimizes the MVD and utilize RGS to construct and streamline the fusion graph $g$ to identify the longest path and delete other shorter paths that may cause conflicts that accuracy is improved in partial cases.

In summary, EPRM constructed in this study and its application in aircraft velocity ranking demonstrate its superiority in handling uncertainty and conflicting information, and provides a new decision making strategy for the field of multi-sensor data fusion. Future research can continue to explore how to expand the application scope of this model in more complex environments and more diverse data types, while improving the fusion algorithm to adapt to higher data quality and real-time requirements. In addition, EPRM needs to be further optimized in order to improve the robustness and accuracy of the decision making system on a larger scale.

\bibliographystyle{IEEEtran}
\bibliography{reference}

\end{document}